%% file: main.tex
\documentclass[Afour,sageh,times]{sagej}

\usepackage{moreverb,url}
\usepackage[normalem]{ulem}
\usepackage[colorlinks,bookmarksopen,bookmarksnumbered,citecolor=red,urlcolor=red]{hyperref}
\usepackage[utf8]{inputenc}
\usepackage{siunitx}
\usepackage[ruled,vlined]{algorithm2e}
\newcommand\BibTeX{{\rmfamily B\kern-.05em \textsc{i\kern-.025em b}\kern-.08em
T\kern-.1667em\lower.7ex\hbox{E}\kern-.125emX}}

\newcommand{\revise}[2]{\textcolor{red}{\sout{#1}}\textcolor{blue}{#2}}

\def\volumeyear{2025}
\def\volumenumber{}     
\def\issuenumber{Under Review} 

\begin{document}

\runninghead{Meng et al.}
\def\journalname{The International Journal of Robotics Research}

\title{Sensor-Space Based Robust Kinematic Control of Redundant Soft Manipulator by Learning}

\author{Yinan Meng\affilnum{1}, Kun Qian\affilnum{1}, Jiong Yang\affilnum{1}, Renbo Su\affilnum{1}, Zhenhong Li\affilnum{2}, \\ and Charlie C.L. Wang\affilnum{1}}

\affiliation{\affilnum{1}Department of Mechanical and Aerospace Engineering, The University of Manchester, Manchester, United Kingdom.\\
\affilnum{2}Department of Electrical and Electronic Engineering, The University of Manchester, Manchester, United Kingdom.}

\corrauth{Charlie C.L. Wang, Dept. of Mechanical and Aerospace Engineering, 4th Floor, Nancy Rothwell Building, Booth St E, Manchester M1 7HF, United Kingdom.}

\email{changling.wang@manchester.ac.uk}

\begin{abstract}
The intrinsic compliance and high degree of freedom (DoF) of redundant soft manipulators facilitate safe interaction and flexible task execution. However, effective kinematic control remains highly challenging, as it must handle deformations caused by unknown external loads and avoid actuator saturation due to improper null-space regulation -- particularly in confined environments. In this paper, we propose a Sensor-Space Imitation Learning Kinematic Control (SS-ILKC) framework to enable robust kinematic control under actuator saturation and restrictive environmental constraints. We employ a dual-learning strategy: a multi-goal sensor-space control framework based on reinforcement learning principle is trained in simulation to develop robust control policies for open spaces, while a generative adversarial imitation learning approach enables effective policy learning from sparse expert demonstrations for confined spaces. To enable zero-shot real-world deployment, a pre-processed sim-to-real transfer mechanism is proposed to mitigate the simulation-to-reality gap and accurately characterize actuator saturation limits. Experimental results demonstrate that our method can effectively control a pneumatically actuated soft manipulator, achieving precise path-following and object manipulation in confined environments under unknown loading conditions.
\end{abstract}


\maketitle


\input{Introduction}
\input{Preliminaries}
\input{Environment}

\input{Reinforcement_Learning}

\input{Reward_Generation}
\input{Experimental_Results}
\input{Conclusion}

\begin{dci}
The author(s) declared no potential conflicts of interest with respect to the research, authorship, and/or publication of this article.
\end{dci}

\begin{funding}
The project is partially supported by the chair professorship fund of the University of Manchester and the research fund of UK Engineering and Physical Sciences Research Council (EPSRC) (Ref.\#: EP/W024985/1).
\end{funding}

\begin{sm}
Supplemental video for this article is available online at: \\\url{https://youtu.be/nIJ1ZCadZB8}.
\end{sm}

\bibliographystyle{SageH}       
\bibliography{IJRR}


\end{document}

%% file: Introduction.tex
\section{Introduction}\label{secIntro}
Soft manipulators have gained increasing attention due to their ability to enhance manipulation in complex and cluttered environments (\cite{rus2015}). However, achieving robust kinematic control of these redundant mechanisms remains challenging due to their motion's strong sensitivity to external loads, actuation limits, and environmental constraints. 

\subsection{Challenges in Redundant Soft Manipulator Control}
\begin{figure}[t]
\includegraphics[width=\linewidth]{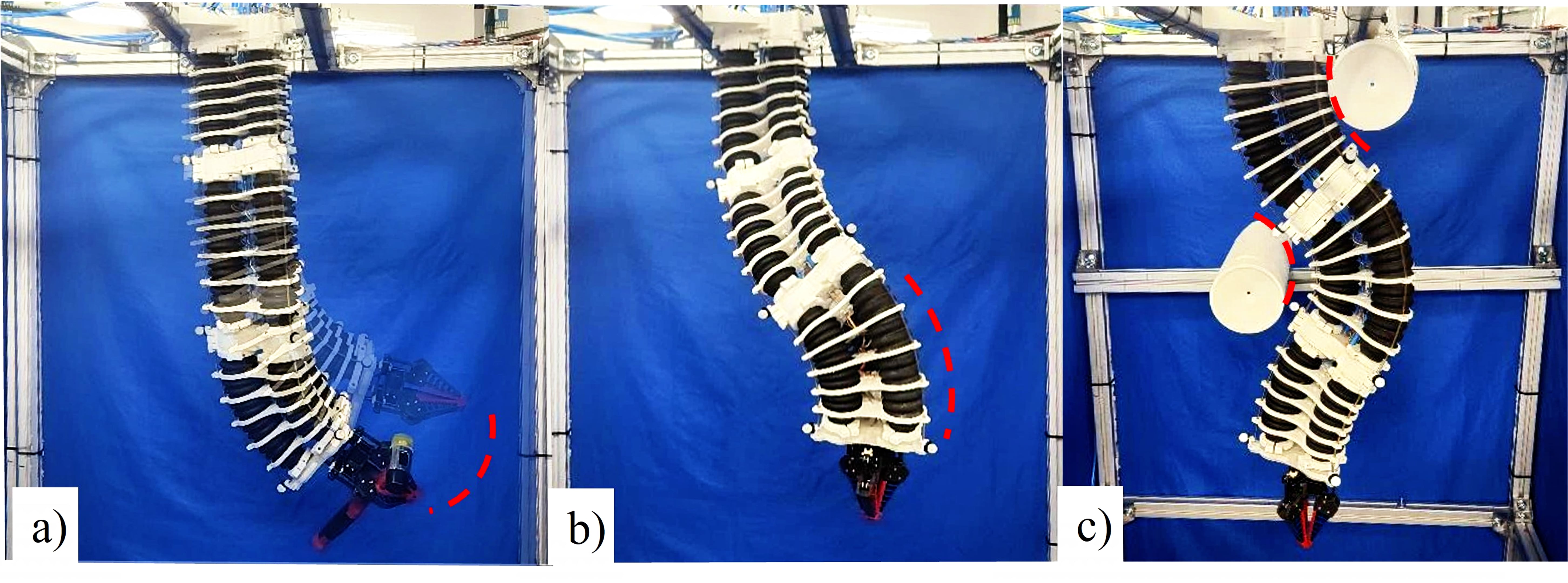}
\caption{Challenges encountered in the kinematic control of the redundant soft manipulator include a) unknown loads, b) actuator saturation and c) operation within confined spaces.}
\label{fig:problem3}
\end{figure}

Load dependency poses a significant challenge in the kinematic control of redundant soft manipulator, as their high compliance leads to passive deformations under external forces, causing inconsistencies in actuation response (\cite{bruder2023}). 
Specifically, the hyperelasticity of their actuators results in a highly nonlinear and load-sensitive actuation-to-configuration mapping (see, Fig.~\ref{fig:problem3}(a)) in manipulation tasks involving large deformations. Model-based kinematic solutions always assume uniform, symmetrical structures in a quasi-static state (e.g., \cite{della20202,della2023,shao2023}), but these assumptions are not satisfied under varying loads, leading to significant control errors when the deformation-induced drifts are not compensated (ref.~\cite{renda2014,armanini2023}). Recent data-driven methods attempt to model the actuation-to-configuration mapping based on non-contact sensor feedback (e.g., \cite{almanzor2023,tang2024}), but they require extensive training data to accommodate variable load-induced deformations, resulting in limited generalizability. 

Actuator saturation also significantly impacts the kinematic control of soft manipulators (see Fig.~\ref{fig:problem3}(b) for an example). In redundant soft continuum manipulator, inverse kinematics (IK) is particularly prone to saturation due to their many-to-one actuation-to-state mapping (\cite{martin2018}), where multiple actuator solutions in actuation space can produce the same end-effector configuration in task-space. Gradient-based methods (e.g., \cite{meng2023}) perform local optimization and are struggle with null-space regulation due to inherently nonconvex actuation-to-state mapping. Sliding mode control (\cite{shao2023model}) handles nonlinear actuator dynamics but causes chattering effects, and leads to inefficient pressure distribution over time due to its lack of trajectory-level optimization. Without proper null-space regulation, these methods can push some actuators to their limits without contributing to the motion of end-effector.

To enable effective manipulation in confined space -- which is in fact the advantage of soft robots than the conventional rigid-body robots, environmental constraints need to be incorporated into kinematic control. Unlike rigid manipulators, soft manipulators require dynamic constraints that adapt to configuration variations. On the other hand, traditional IK methods with hard constraints can overly restrict compliance, preventing the manipulator from leveraging its deformability to reach the goal. A common approach to addressing these issues is incorporating task-space constraints into configuration optimization (\cite{cao2022,lai2022}). However, this often leads to overly conservative IK solutions, inaccurate task execution, and increased computational complexity, especially in redundant setups. As illustrated in Fig.~\ref{fig:problem3}(c), effective operation in confined spaces demands both feasible IK solutions and precise task execution, posing additional challenges for soft manipulator control.

This paper presents a Sensor-Space Imitation Learning-based Kinematic Control (SS-ILKC) framework for redundant soft manipulators, which can effectively address the aforementioned challenges of unknown loads, actuation saturation, and operations in confined-spaces. By using the scattered sample poses demonstrated by experts in confine spaces and the training dataset generated in simulation for open spaces, SS-ILKC enables robust IK solutions for emulating demonstrated motions. Additionally, a sim-to-real (S2R) transfer mechanism is integrated for accurate and zero-shot deployment of control policies learned from simulation. Experimental tests have been conducted to demonstrate the robustness of SS-ILKC in manipulating objects of varying weights for path-following tasks and pick-and-place operations in confined spaces.

\subsection{Related Work}
The conventional approach to the kinematic control of soft manipulators involves modeling their kinematics and dynamics, followed by designing model-based feedforward and feedback controllers for quasi-static point reaching (\cite{till2019,della20201,fang2020}). To address unknown loads that cause large deformations in the robot configuration space, \cite{bruder2021} proposed a model predictive control scheme with a Koopman operator-based loading parametrization. \cite{huang2021} introduced a variable curvature modeling approach that maps the relation between the actuation space and the configuration space while accounting for different payloads. An alternative approach leverages external sensors with data-driven methods to directly regulate the robot’s configuration. For instance, \cite{zhao2021} used strain gauge sensor and trained an LSTM neural network with configuration torques and external payload data to predict key actuation points. Moreover, \cite{fang2019} proposed an eye-in-hand visual servo method to resolve nonlinear IK solution under external loads by Gaussian regression-based image feature mapping. However, dependence on vision hardware and the complex calibration process (for both the camera and the actuation-to-configuration mapping) limit their applicability in scenarios where the camera is obstructed, the lighting condition is poor, or the manipulator undergoes large deformations that distort visual feedback. 

For rigid robots, actuation saturation can be typically managed by incorporating actuation dynamics into the controller design (\cite{yang2020, chen2024}). However, these methods are less effective for soft manipulators due to their state-dependent actuation dynamics that are difficult to be modeled. Recent studies have explored reinforcement learning (RL) for soft robot control, which eliminates the need for actuation modeling and allows the incorporation of control limits as hard constraints during data-driven optimization. \cite{thuruthel2018} introduced a model-based RL method for handling payload variations via progressive point-to-point reaching. \cite{centurelli2022} integrated deep RL with an LSTM-NN forward kinematic model, enabling efficient path-following under varying loading conditions. Recent RL-based soft robot controllers overcomed actuation limits for complex tasks but rely on external sensors and additional actuators. For instance, \cite{jitosho2023} modeled a soft robot as an n-link pendulum with hinge joints, using external linear actuators to train ``swing" maneuvers. Similarly, a force sensor-based dynamic control policy was learned for robot ``pushing" tasks (\cite{alessi2024}). However, enforcing hard actuation limits can lead to diminished controllability, limited stability margins, and reduced robustness to external disturbances and real-world uncertainties. Moreover, the lack of saturation representation in simulated environments oversimplifies actuation-to-configuration mapping, leading to discrepancies between simulated training and real-world deployment. 

A key advantage of soft manipulator is its ability to perform flexible manipulation in complex and cluttered environments, going beyond standard task-space reaching by adapting to highly constrained robot configuration. To enable soft robot operation in free space with obstacles, \cite{lai2022} modeled axial compression and segment coupling using the compressible curvature approach for null-space control via an optimization-based kinematic solver. Similarly, \cite{cao2022} developed a collision map to translate task-space obstacles into configuration space constraints, enabling efficient path following. Several studies leverage vision-based approaches that use image feedback for task guidance. For instance, \cite{wang2020} proposed a hybrid visual servoing controller integrating a contact localization algorithm that decouples static and movable robot segments for repositioning contacts in task-oriented settings. Combined with tactile feedback, compliant obstacle avoidance was achieved using a control barrier function with safe interaction constraints  (\cite{fan2024}). Recent work (\cite{nazeer2024}) introduced two control strategies for successfully transferring a trained RL policy to manage various task-space obstacles, even in the presence of actuation failure. 
However, operating redundant soft manipulators in confined spaces (such as T-pipes demonstrated later in our paper) introduces additional challenges, including kinematic complexities arising from redundant degrees of freedom (DoFs) and collisions with environmental constraints in the configuration space to avoid contact-induced deformation.
Currently, no off-the-shelf solutions exist to effectively address such challenging tasks.

\subsection{Our Method}
\begin{figure}[t]
\centering
\includegraphics[width=\linewidth]{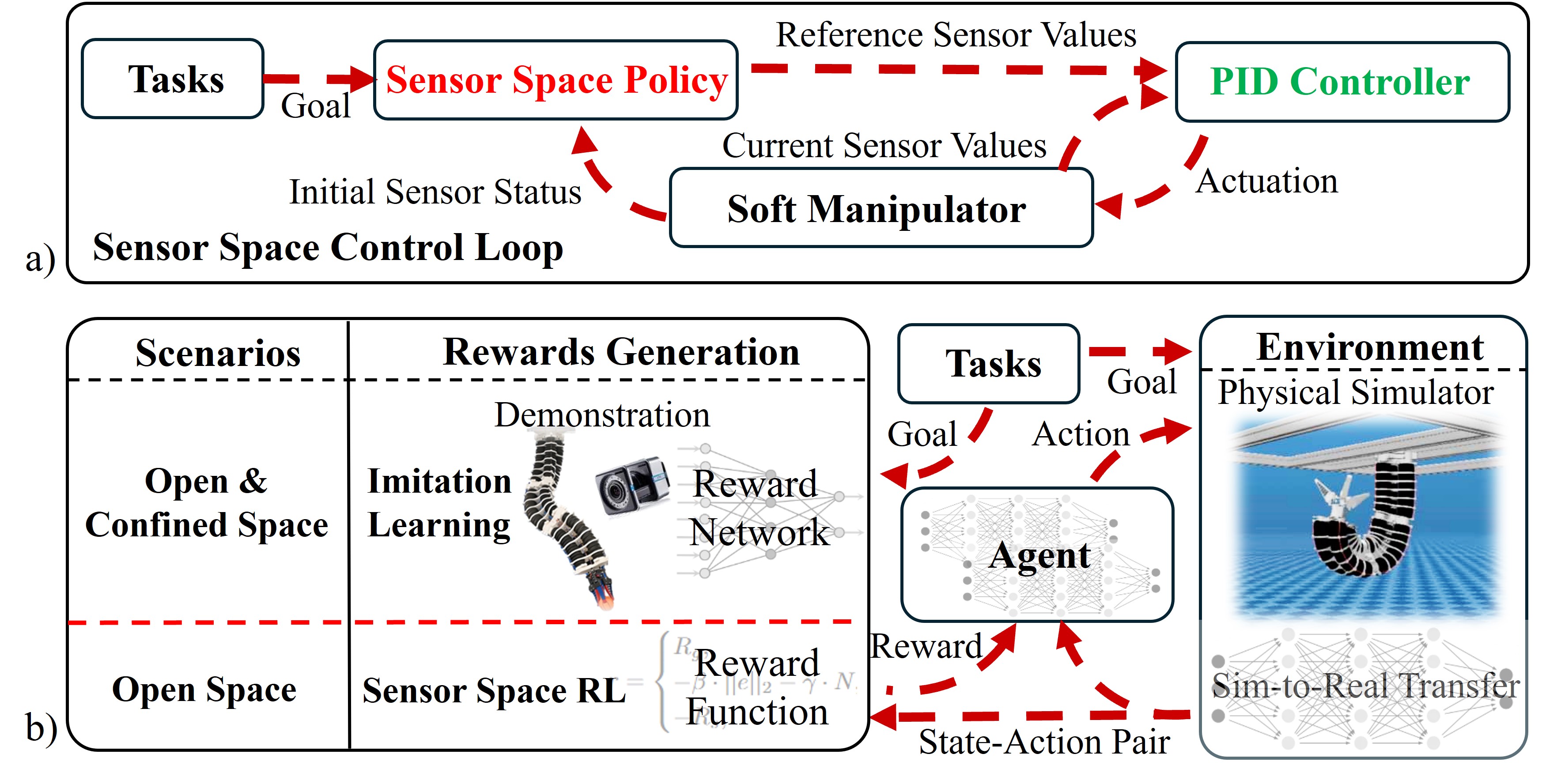}
\caption{Schematic diagrams of (a) the sensor-space based control loop and (b) our SS-ILKC framework learning the control policies from both simulation (for open space) and physical demonstration (for confined space) -- i.e., imitation learning.
}\label{fig:overall_loop}
\end{figure}

Building upon our prior work in sensor-space control (\cite{meng2023}), we employ spring-based sensors to directly observe the manipulator’s geometric state, thereby defining the control objective in terms of its geometry (i.e., spring length) rather than actuation attributes that are highly sensitive to varying load conditions. To effectively mitigate actuator saturation, a learning-based strategy is adopted to develop sensor-space control policies using data-driven methods. As illustrated in Fig.~\ref{fig:overall_loop}(a), the sensor-space policy takes the initial sensor status and the task goal as inputs, then outputs reference sensor signals to guide the manipulator’s actuation. A PID-based controller is subsequently employed to track these reference signals in real time, ensuring accurate manipulation. This control method is load-independent and can effectively address actuator saturation. In our implementation, the control objectives are defined as 6D end-effector poses for robot-reaching tasks.

Our approach adopts a reinforcement learning (RL)-inspired framework for control policy learning, capitalizing on RL's ability to explore complex environments and generate adaptive control strategies. While model-free RL avoids the need for precise kinematic models, applying it to redundant soft manipulators remains non-trivial due to challenges such as workspace-wide generalization (i.e., goal-conditioned control discussed in \cite{schaul2015universal,kaelbling1993learning}), sample inefficiency, and difficulty in achieving precise control under high nonlinearity and redundancy. To overcome these limitations, we introduce a multi-goal, sensor-space RL framework trained in a high-fidelity simulation environment that:
\begin{itemize}
\item Implements sensor-space control to ensure load-independent policy behavior, while explicitly incorporating actuation constraints into the optimization process;

\item Conducts multi-goal learning to enable generalization across the entire workspace, establishing robust mappings from diverse initial configurations to arbitrary target goals.
\end{itemize}
A central component of our method is a pre-processed sim-to-real (S2R) transfer mechanism, designed to correct kinematic discrepancies between simulation and the physical system. In contrast to conventional approaches that adapt policies post-training using real-world data -- e.g., via the fine-tuning method presented in (\cite{hwangbo2019learning,lee2019robust}), our method performs pre-calibration by training a lightweight neural network to map simulated end-effector poses to their real-world counterparts. This pre-calibrated simulator enables policy learning under realistic actuator saturation and pose constraints. The proposed S2R stage provides:
\begin{itemize}
\item Accurate estimation of saturation limits and physical parameters through offline optimization;
\item A simulation environment corrected for physical fidelity, facilitating effective policy training;
\item Elimination of the need for post-training fine-tuning, ensuring direct transferability of learned policies.
\end{itemize}
By embedding this S2R calibration step into policy learning, the trained RL agent can be employed to zero-shot deployment without online adaptation.

Furthermore, conventional RL methods often struggle to learn effective policies for controlling soft manipulators in confined spaces. This is primarily due to the extensive prior knowledge required for reward function design and the challenges of accurately replicating complex environmental interactions in simulation for training data generation. To overcome these limitations, we incorporate generative adversarial imitation learning (GAIL) in our framework, enabling the RL agent to effectively learn control policies from sparse expert demonstrations. As illustrated in Fig.~\ref{fig:overall_loop}(b), our SS-ILKC framework integrates IL-based reward generation, sensor-space RL, and a dedicated simulation environment with S2R transfer to achieve load-independent and saturation-free operation in both open and confined spaces.

\subsection{Contributions}
The main contributions of this paper are summarized as follows:
\begin{itemize}
%
%
\item A novel learning-based framework is presented to enable sensor-space based robust control of highly redundant soft manipulators by learning control policies from both simulation (for open space) and physical demonstration (for confined space). 

\item A multi-goal sensor-space learning method is proposed to derive load-independent, saturation-free control policies with zero-shot real-world deployment and workspace-wide generalization, which is inspired by RL principles and integrated with a pre-processed sim-to-real mechanism.

\item A GAIL-based demonstration learning module is developed to enable robust policy optimization in confined spaces, complementing simulation-based training taken in open environments.

\end{itemize}
By our learning-based sensor-space control framework, we are able to address three key challenges in soft manipulator's control, i.e., unknown loads, actuator saturation and confined-space operations. The effectiveness of this approach has been demonstrated on a physical setup of pneumatically actuated soft manipulator for different tasks. 

This paper is structured as follows. Sec.~\ref{Preliminaries} presents the soft manipulator hardware and sensor integration. Sec.~\ref{secSim} introduces the simulation environment with S2R transfer which facilitates the training of the RL algorithm presented in Sec.~\ref{secRL}. The proposed SS-ILKC framework, along with detailed algorithmic formulations for GAIL-based reward learning, is described in Sec.~\ref{secIL}. Experimental results validating the robustness and generalizability of SS-ILKC are reported in Sec.~\ref{secResult}. Finally, Sec.~\ref{secConclusion} concludes the paper and discusses possible future works.

%% file: Preliminaries.tex
\section{Structural Design and Sensor Integration}\label{Preliminaries}
\begin{figure}[t]
\includegraphics[width=\linewidth]{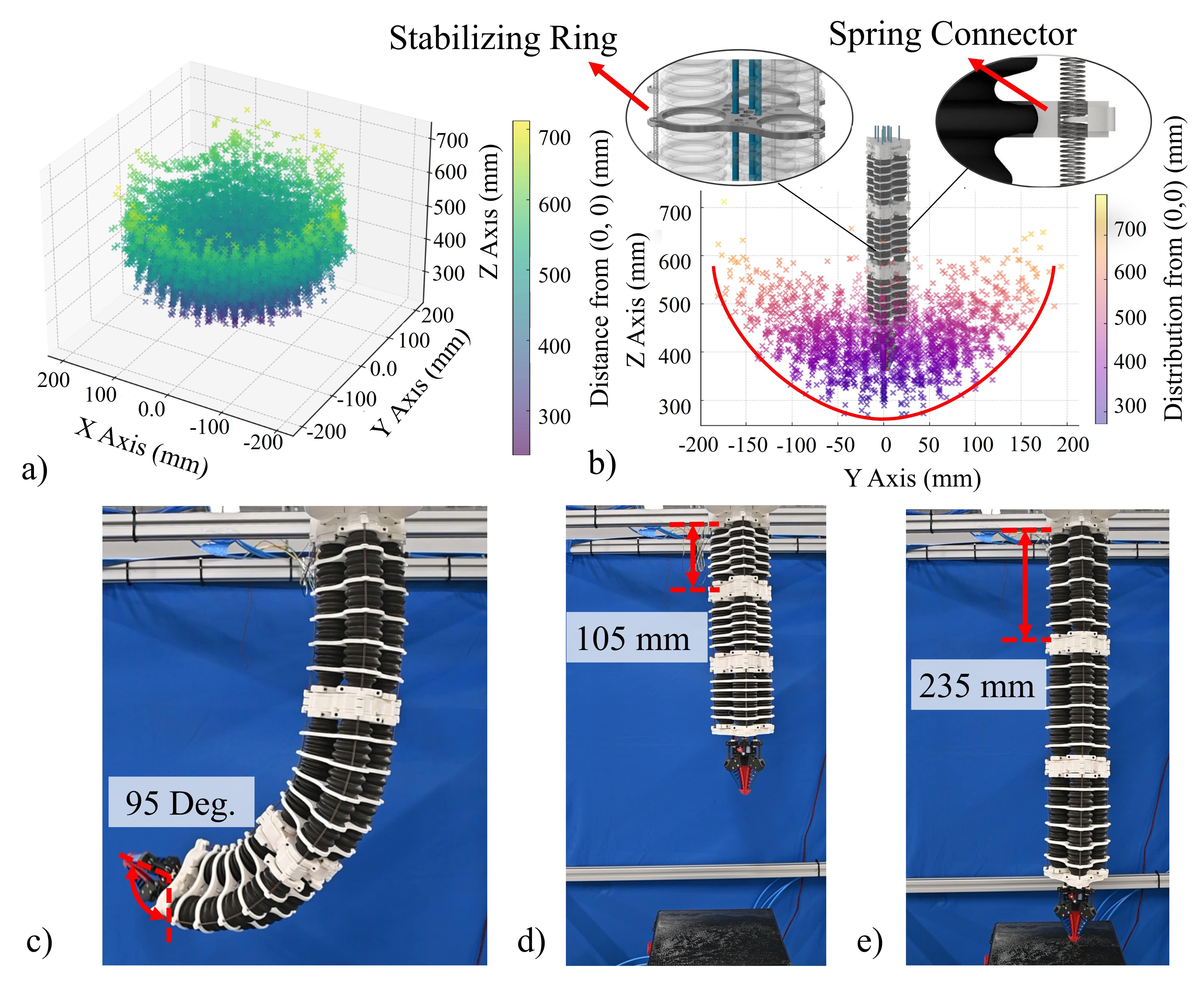}
\caption{Workspace of the redundant soft manipulator. (a) A 3D representation of the workspace, with the color gradient indicating the distance from the origin. (b) A 2D projection of the workspace on the YZ-plane, with the red curve representing the theoretical boundary of the workspace determined by the manipulator's actuation limits. The insets illustrate the stabilizing rings and the integrated conductive springs, which gives a shape-conforming design. (c) Maximum bending angle of the soft manipulator can be \SI{95}{\degree}, (d) the minimum length of one section is \SI{105}{mm}, and (e) the maximum length of one section is \SI{235}{mm}.
}\label{fig:workspace}
\end{figure}
This section describes the hardware design of the soft manipulator used in our work, as shown in Fig.~\ref{fig:workspace}. The manipulator is a pneumatically actuated soft continuum arm composed of three modules, each integrated with spring-shaped inductive sensors that facilitate sensor-space-based kinematic control.

\subsection{Hardware of Soft Manipulator}
The soft manipulator is a three-section pneumatically actuated continuum manipulator, with each section containing three parallel chambers (V6-00325 from Freudenberg) featuring a wall thickness of \SI{2}{mm}, an inner diameter of \SI{20}{mm}, an outer diameter of \SI{59}{mm}, and a free length of \SI{200}{mm}. The chambers are actuated by air inflation and deflation, resulting in longitudinal extension and compression. Stabilizing rings are placed on every two chamber ridges to prevent irregular buckling, ensuring pressure is directed toward longitudinal motion. Each ring is designed with round holes for securely attaching conductive springs, which serve as shape-conforming sensors that vary with chamber deformations, as shown in the insets of Fig. \ref{fig:workspace}. A lightweight gripper (\SI{0.45}{kg}) is mounted on a 3D-printed base to serve as the end-effector. Together with the gripper, the total weight of the manipulator is approximately \SI{4.7}{kg}.

Each chamber is controlled by a YOUCHEN VN-C1 brushless pump unit at an airflow rate of \SI{12}{L/min}. Each pump is equipped with two 3-way, 2-position valves that switch between inflation and deflation within \SI{10}{ms}, supporting real-time control up to \SI{50}{Hz}. To determine the reachable workspace, each chamber is systematically actuated at three pressure levels  as (\SI{-40}{kPa}, \SI{0}{kPa}, and \SI{20}{kPa}). The range of reachable points are visualized in Fig.~\ref{fig:workspace}(a), with a color gradient indicating the distance from the origin. The YZ-plane projection in Fig. \ref{fig:workspace}(b) shows a vertical coverage of \SI{350}{mm} and a lateral coverage of \SI{400}{mm}, with the boundary of workspace depicted by the red curve. Besides, Figs.~\ref{fig:workspace}(c)–(e) illustrate maximum bending angles of the soft manipulator up to \SI{95}{\degree} and vertical length variations of each section ranged from \SI{105}{mm} to \SI{235}{mm}.

\subsection{Sensor Integration and Signal Processing}
The soft manipulator utilizes springs as sensors to provide feedback on chamber deformation. These springs are made of piano steel, with a wire diameter of \SI{0.4}{mm}, an outer diameter of \SI{4}{mm}, a pitch of \SI{0.4}{mm}, and resulting in a force rate of \SI{0.054}{N/mm}. The small force rate ensures that the springs minimally affect the soft manipulator's movement. At the boundary of each module, three springs are securely attached to the stabilizing rings, ensuring they deform in sync with the chambers. We use the LDC1612 (Texas Instruments) as an inductance-to-digital converter to capture sensor variations through the resonant frequency $f_{\text{sensor}}$, which is denoted as the sensor feedback. The inductance is expressed as  
\begin{align}\label{freq_to_ind}
    I_{\text{spring}} = \frac{1}{C \cdot (2\pi f_{\text{sensor}})^2} - I_{\text{comp}},
\end{align}  
where $C$ is the total capacitance in the resonant circuit, \( f_{\text{sensor}} \) is the resonant frequency, and $I_{\text{comp}}$ = \SI{1000}{pF} is the inductance offset. 
Moreover, the inductance of the conductive spring is related to its length as
\begin{align}\label{lenght_to_ind}
I_{\text{spring}} = \frac{\mu_{p} \mu_0 N^2 \pi r_{p}^2}{l},
\end{align}  
where $I_{\text{spring}}$ is the inductance of the spring, $\mu_{p}$ is the relative permeability of the core (dimensionless), $\mu_{0}$ is the permeability of free space, $N$ is the number of turns, $r_{s}$ is the spring radius, and $l$ is the spring length. Since the spring has no core material, its permeability is set to air, $\mu_{0} = 1.26 \times 10^{-6}$, making $\mu_{p} \approx 1$. 
By combining Eqs.\eqref{freq_to_ind} and \eqref{lenght_to_ind}, the relationship between the sensor feedback $f_{\text{sensor}}$ and spring length $l$ is derived as  
\begin{align}\label{freq_to_length}
l = \frac{\mu_0 N^2 \pi r_{p}^2 C (2\pi f_{\text{sensor}})^2}{1 - C I_{\text{comp}} (2\pi f_{\text{sensor}})^2}.
\end{align}
This shows that the sensors can directly measure the soft manipulation's configurations as the lengths of springs by the resonant frequency $f_{\text{sensor}}$, which enables the sensor-space based control.


%% file: Environment.tex
\section{Simulation and Sim-to-Real for Learning}\label{secSim}
Our SS-ILKC framework learns effective control policies through trial-and-error operations. To effectively generate datasets for training, we first employ a high-fidelity simulation environment based on Multi-Joint dynamics with Contact platform (MuJoCo). To bridge the gap between simulation and reality, we introduce a sim-to-real transfer mechanism that enables accurate deployment of simulation-trained policies in a zero-shot manner and also facilitates the transformation of realistic demonstrations back into effective training samples. 
\begin{figure}[!t]
\centering
\includegraphics[width=\linewidth]{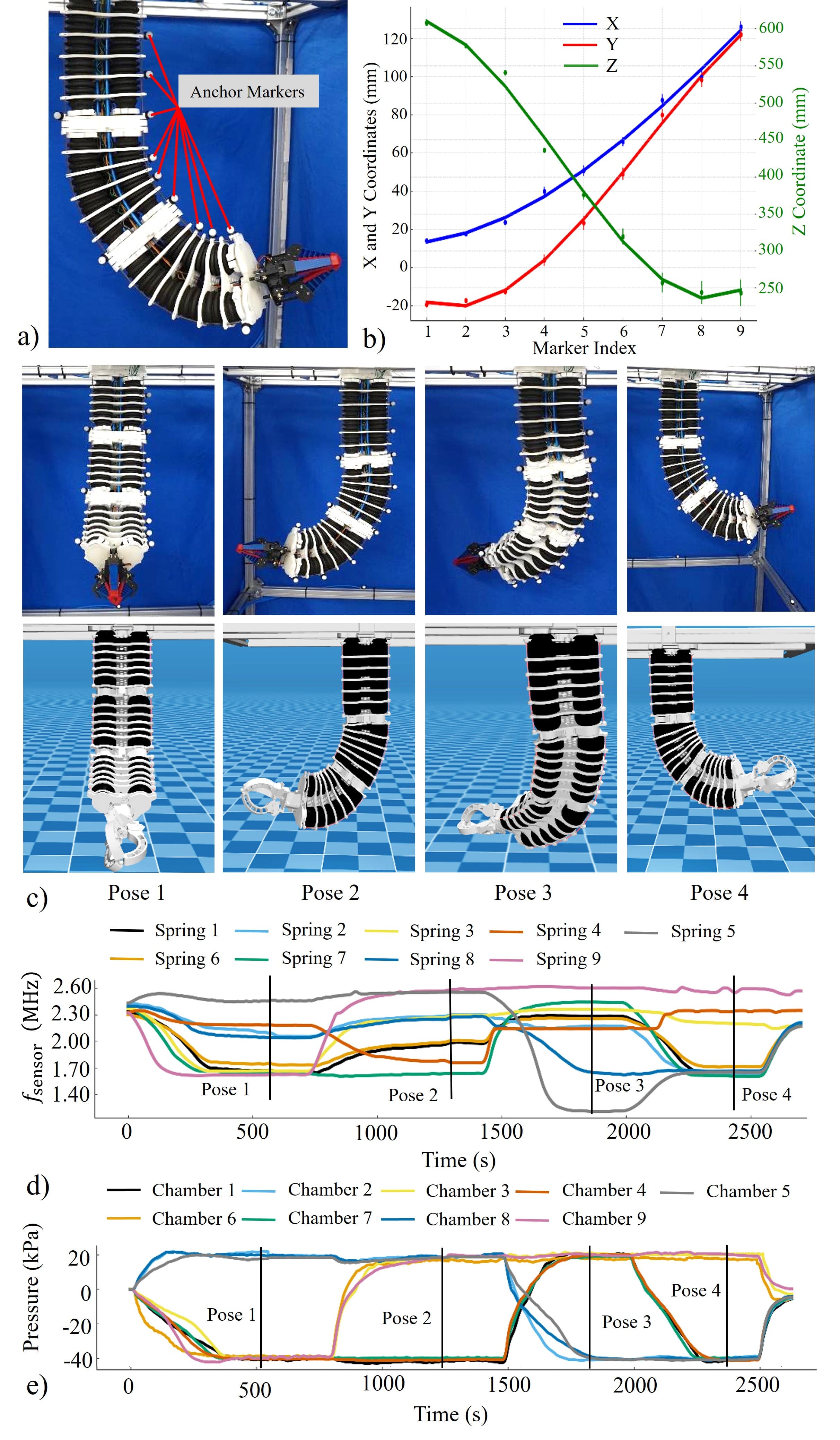}
\caption{Soft manipulator calibration procedure and results: 
(a) The markers on soft manipulator are employed to obtain its configuration by the motion capture system, 
(b) polynomial fitting of X, Y, and Z coordinates for markers under maximum bending, 
(c) comparison of simulated and real manipulator poses in various configurations,  
(d) spring frequency response corresponding to different poses, and 
(e) recorded actuator pressures for different poses.}
\label{fig:Sim_comp}
\end{figure}

\subsection{Simulation Environment}
For existing off-the-shelf simulators (e.g.,~\cite{goury2018fast,van2019spatial,hu2019chainqueen}), they usually lack of native support for different actuation methods of soft manipulator such as cable-, tendon-, and pneumatic-driven actuation. By employing a sensor-space control architecture that decodes the soft manipulator’s configuration via spring geometries, our approach eliminates the need of explicit actuator modeling while maintaining accurate static behavior in the MuJoCo (\cite{todorov2012mujoco}) simulation environment (as illustrated in Fig.~\ref{fig:overall_loop}). 

Specifically, we model pneumatic chambers in the simulator using nine tendons with a diameter of \SI{60}{mm}, replicating their material properties. Stabilizing rings are placed to align with their real-world counterparts on the soft manipulator. Each fixing plate pair includes two perpendicular revolute joints for out-of-plane rotation, while a cylinder joint allows pure extension and compression, accurately replicating manipulator flexibility. Actuation dynamics are simulated by pneumatic actuators that compress and elongate the tendons. The simulator models spring sensors using nine anchored tendons with an outer diameter of \SI{4}{mm}. The tendons' stiffness and damping coefficients are defined as \SI{0.054}{N/mm} and \SI{0.03}{Ns/mm}, respectively. Since the simulator cannot replicate the realistic inductive sensor feedback \( f_{\mathrm{sensor}} \), nine spring length variations \( L_{\mathrm{spring}}=[l^n] \in \mathbb{R}^9 \), where \( n = {1, 2, \ldots, 9} \), are employed to conduct sensor-space control in the simulation environment. The lengths can also be converted back into $f_{\mathrm{sensor}}$ as will be detailed in Sec.~\ref{subsec:S2R}. With these settings in MuJoCo, the simulator replicates the soft material’s response under pneumatic actuation, creating a realistic soft manipulator simulation environment.

\subsection{Simulation Calibration and Validation}
\label{subsection_calibration}
To effectively model actuator saturation, the simulator is expected to precisely replicate the physically constrained poses of the soft manipulator. 
We conduct a calibration process by a Motion capture system to ensure this.
Fig.~\ref{fig:Sim_comp}(a) demonstrates the calibration setup, with nine markers attached along one side of the soft manipulator. Locations of markers are collected at different poses with large bending, with the corresponding spring signals and pressure values shown in Figs.~\ref{fig:Sim_comp}(d) and (e). 

The spatial locations of anchor markers obtained on Pose 1 -- i.e., the one with maximum bending are then fitted with three independent forth-order polynomial functions, corresponding to the X, Y, and Z coordinates as shown in Fig. \ref{fig:Sim_comp}(b).
These polynomial functions are applied as length constraints of the tendons in the simulator for modeling their deformation. Subsequently, the stiffness and damping parameters of the chambers are adjusted accordingly so that the simulated end-effector poses closely align with real-world counterparts. The calibration results are illustrated in Fig.~\ref{fig:Sim_comp}(c), showing a close match between the soft manipulator’s configuration captured in simulation and that observed in the real-world.

\begin{figure}[t]
\centering
\includegraphics[width=\linewidth]{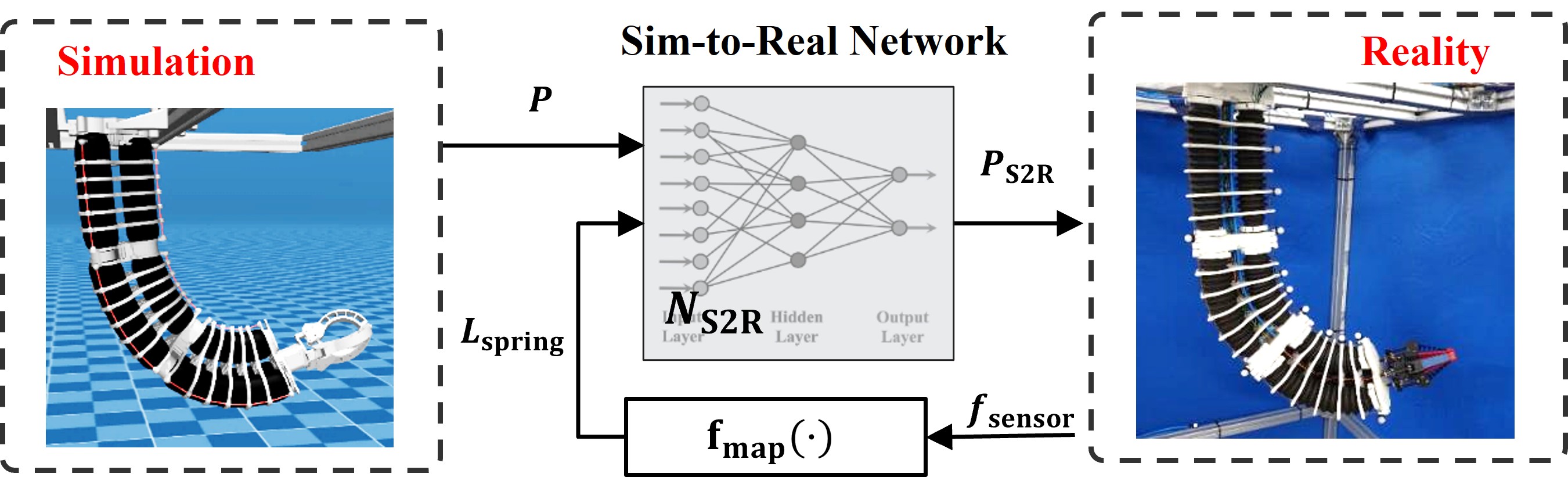}
\caption{The proposed Sim-to-Real (S2R) transfer employs a neural network bridges the gap between simulation and reality.}
\label{fig:S2R}
\end{figure}

\subsection{Sim-to-Real Transfer}\label{subsec:S2R}
Although calibration process has been applied in Sec. \ref{subsection_calibration} to improve the fidelity of the simulation, the discrepancies between the simulation and reality persist, e.g., unmodeled hysteresis of actuators, frictions, etc. To further mitigate the sim-to-real gap, we introduce a pre-processed S2R mechanism (executed before control policy training) to align the simulated and real end effector pose based on real-world demonstrations, i.e.,  which implicitly captures the actuator saturation boundary. Due to the high fidelity of the S2R enhanced simulation, our approach enables zero-shot deployment of the learned policy without post-processing fine-tuning. 
Specifically, we propose a lightweight neural network to align simulated and reality through the following steps (Fig.~\ref{fig:S2R}):
\begin{enumerate}
    \item Collect sensor value $f_{\mathrm{spring}}$ which drives the soft manipulator’s motion and the corresponding real-world end-effector poses;
    
    \item Determine the corresponding spring length \(L_{\mathrm{spring}}\) in simulation using the mapping function \(\mathbf{f}_{\mathrm{map}}(\cdot)\);
    
    \item Conduct sensor-space control in the simulation by using \(L_{\mathrm{spring}}\) as reference and obtain the simulated end-effector pose \(P \in \mathbb{R}^{6}\);
    
    \item Train the S2R network \(N_{\mathrm{S2R}}\), using \((P, L_{\mathrm{spring}} )\) as input and \(P_{\mathrm{S2R}}\in \mathbb{R}^{6}\) as output.
\end{enumerate}
For \(N_{\mathrm{S2R}}\), the mean squared error is utilized as the loss function to facilitate training, while the Adam optimizer is employed with a learning rate of \(0.01\) and a batch size of \(128\). For the mapping function \( \mathbf{f}_{\mathrm{map}}(\cdot) \), the spring lengths in the simulation environment and the real world are first aligned through a linear transformation, that is:
\begin{equation}
    L_{\mathrm{spring}} = \beta \cdot l + \delta,
\end{equation}
where \(\beta = 0.93\) and \(\delta = 2.45\) are introduced to compensate for manufacturing variability, including fabrication tolerances and sensor alignment errors. Based on (\ref{freq_to_length}), the relationship between the $L_{\mathrm{spring}}$ and $f_{\mathrm{sensor}}$ can then be written as:
\begin{equation}
    L_{\mathrm{spring}} = \mathbf{f}_{\mathrm{map}}(f_{\mathrm{sensor}}) = \frac{4 \pi^4 C N^2 \beta f_{\mathrm{sensor}}^2 \mu_0 r^2}{1 - 4 \pi^2 C I_{\mathrm{comp}} f_{\mathrm{sensor}}^2} + \delta.
\end{equation}

By incorporating the S2R transfer mechanism in our policy-based kinematic control framework, the set of simulation results is transformed to construct the action and state for the learning procedure, where \( P_{\mathrm{S2R}} = N_{\mathrm{S2R}}(P, L_{\mathrm{spring}}) \). In conclusion, the proposed simulation environment with S2R mechanism not only accurately model the physical dynamics and actuation constraints of the soft manipulator, but also compensate for unmodeled real-world factors such as dynamic uncertainties and noise. Notably, it also offers high computational efficiency, with each discrete environment execution step completed in \SI{0.03}{s} and each policy training update taking approximately \SI{0.4}{s} to \SI{0.8}{s}.

%% file: Reinforcement_Learning.tex
\section{Sensor Space Control Policy Learning}\label{secRL}

\begin{figure}[t]
\centering
\includegraphics[width=\linewidth]{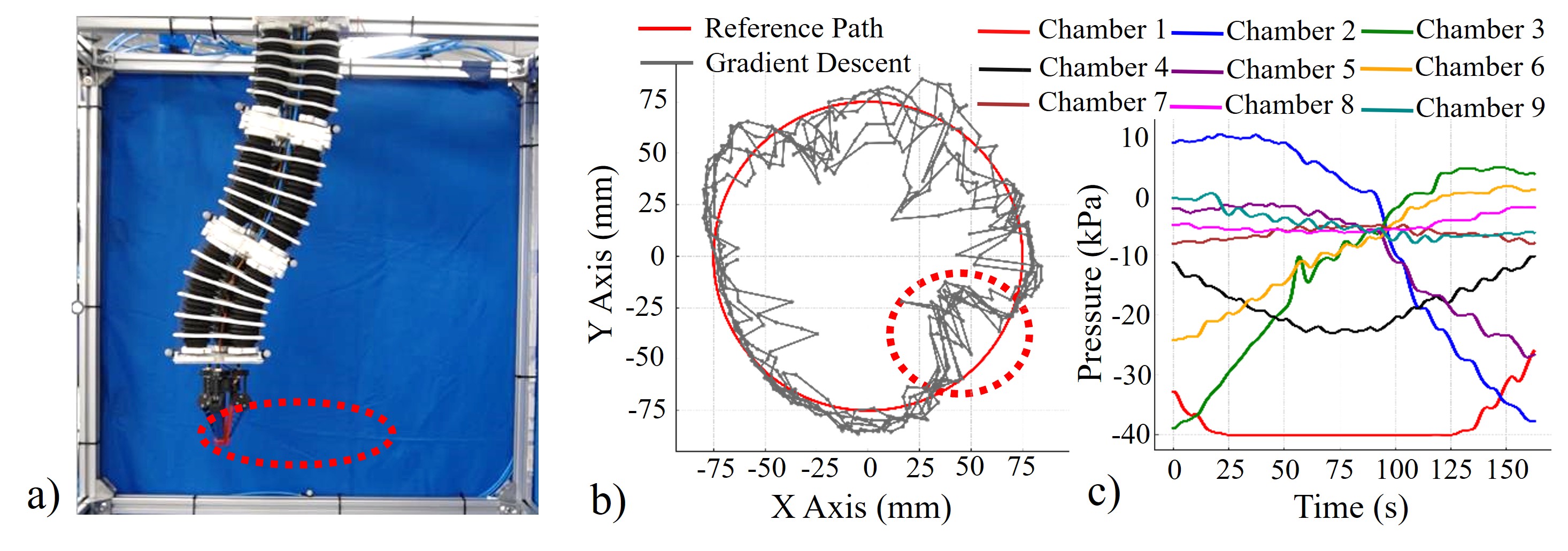}
\caption{The gradient decent method (\cite{meng2023}) lead to actuator saturation in our redundant soft manipulator. (a) Reference path. (b) Actual path significantly deviates from the reference. (c) Control saturation occurs in chamber 1 and leads to path tracking error highlighted in red dotted circle in (b).}
\label{fig:problemGD}
\end{figure}

In our previous work (\cite{meng2023}), a sensor-space controller is developed for load-independent soft manipulator operation in open spaces and the convergence of the proposed gradient descent-based (GD) IK solver heavily relies on the proper selection of the initial guess. As the soft manipulator’s DoFs increase, selecting an appropriate initial guess becomes challenging and will affect the control performance. To further examine this limitation, the GD-based approach was implemented on the redundant soft manipulator, and the results are shown in Fig. \ref{fig:problemGD}. The results indicate that due to the increased complexity, the GD-based IK solver will lead to saturation in some actuators, giving inaccurate kinematic control. To address this issue, we present a sensor-space control framework based on multi-goal RL strategy, trained within our high-fidelity simulation environment and the preprocessed S2R introduced in Sec.~\ref{secSim}
that enables saturation-free control policy optimization.  

\subsection{Definition of Sensor-Space Reinforcement Learning}

Define the desired end-effector pose as the control goal \( g \in \mathbb{R}^{6} \), which comprises three translations and three rotations. In our framework, we aim to achieve robot goal-reaching by following a sequence of states for \( t = 0, \dots, T \), where each step \( t \) corresponds to a quasi-static state and \( T \) denotes the total number of states. The soft manipulator kinematic control problem is then formulated as a Markov Decision Process (MDP), which consists of the 5-tuple \( (\mathcal{S}, \mathcal{A}, \mathcal{P}, r, \gamma) \), where:
\begin{itemize}
    \item $\mathcal{S}$ is the \emph{state space}, where each state $s_t \in \mathcal{S}$ represents the observation and goal condition derived from the soft manipulator sensor space.
    
    \item $\mathcal{A}$ is the \emph{action space}, where actions $a_t \in \mathcal{A}$ are defined as $f_{\mathrm{sensor}}$, which tunes the pneumatic actuation of the nine chambers to realize the desired configuration.
    
    \item $\mathcal{P}(s_{t+1}|s_t, a_t)$ is the \emph{transition probability} from state $s_t$ to $s_{t+1}$ when action $a_t$ is taken.
    
    \item $r$ is the \emph{reward function}, which maps a transition $(s_t, a_t, s_{t+1}, g)$ to a scalar reward signal that evaluates goal-reaching performance. As multi-goal reaching tasks are expected, both the policy and the reward function are conditioned on the given goal. 
    \item $\gamma \in [0, 1)$ is the \emph{discount factor}, discounting future rewards over time.
\end{itemize}

\noindent The objective is to find a control policy \( \pi_{\phi} (a|s) \) parameterized by \( \phi \) that maps states \( s \) to a probability distribution over actions \( a \), and in order to maximize the expected return within the MDP which is given by  
\begin{equation}
    \eta(\pi_{\phi}) = \mathbb{E}_{\tau \sim \pi_{\phi}}\left[ \sum_{t=0}^{T} \gamma^t r(s_t, a_t, s_{t+1}, g) \right].
\end{equation}

\noindent By incorporating saturation-related terms into the reward function design, we aim to train an RL agent that can effectively derive feasible kinematic control policy for the soft manipulator while not violating actuation limits for reaching $g$.

\subsection{Multi-goal Reinforcement Learning Strategy}
\begin{figure}[t]
\centering
\includegraphics[width=\linewidth]{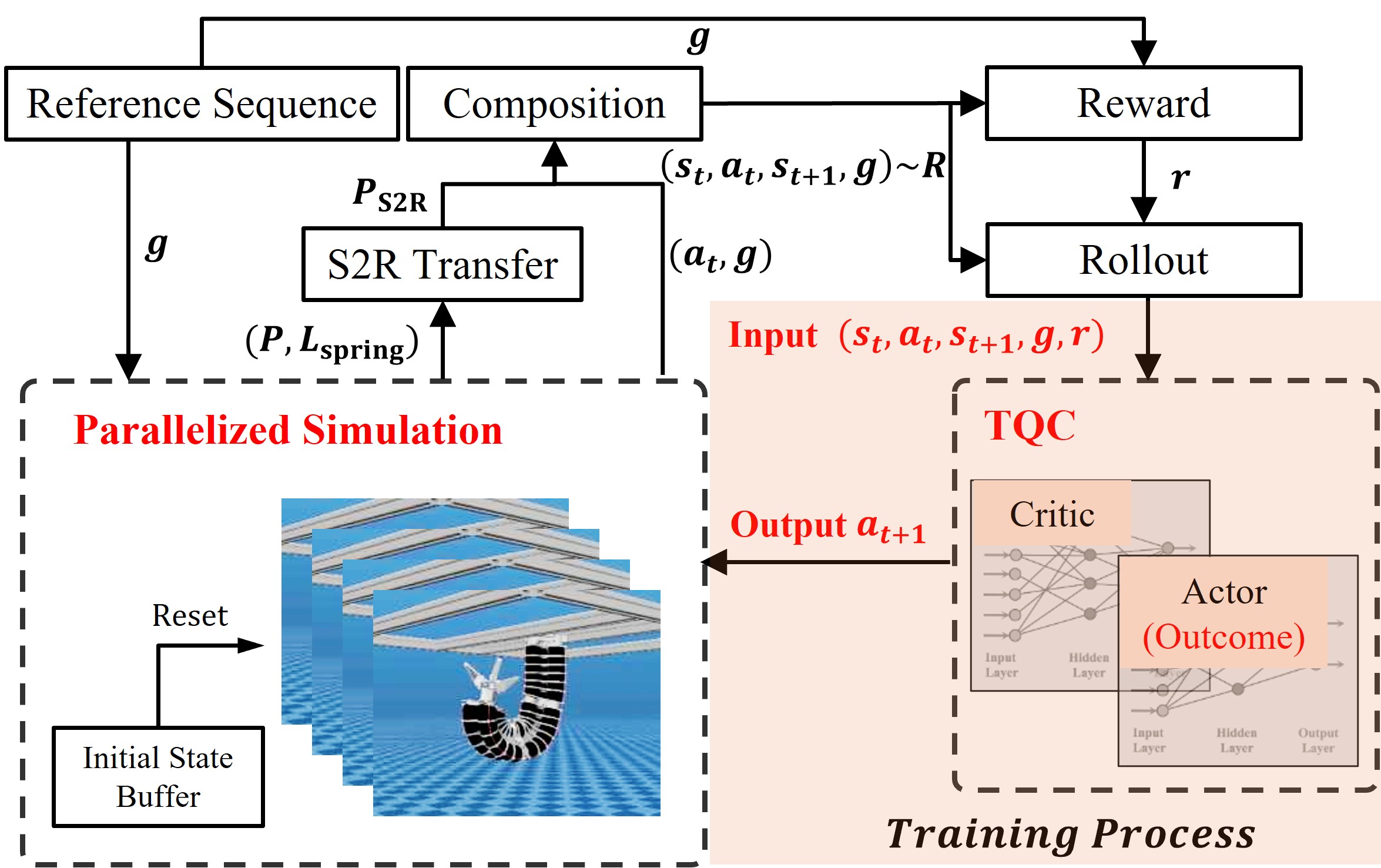}
\caption{Schematic flow of the sensor-space multi-goal RL algorithm for obtaining the soft manipulator's kinematic control policy in simulation.}
\label{fig:RLLoop}
\end{figure}

Recent research has demonstrated the effectiveness of RL for soft manipulator control, particularly in solving IK in high-dimensional redundant spaces (\cite{thuruthel2018,centurelli2022}). However, a robust kinematic control policy for goal-reaching tasks is expected to generalize across the entire robot workspace, ensuring accurate tracking throughout the desired path. Inspired by \cite{plappert2018multi}, which introduced a multi-goal RL algorithm for various tasks such as robot reaching, pushing, and pick-and-place, we introduce a multi-goal RL framework that enables a single optimized policy to generalize across various soft manipulator reaching goals. By utilizing the high-fidelity simulation environment presented in Sec.~\ref{secSim}, actuator saturation attributes are incorporated in the reward function to enhance the robustness of the control policy. Moreover, this sensor-space RL framework serves as the fundamental policy generator for the development of the SS-ILKC framework, which enables confined space operation and will be detailed in Sec.~\ref{secIL}. The schematic of the sensor-space multi-goal RL algorithm can be found in Fig.~\ref{fig:RLLoop}. 

\subsubsection{Elements of RL}
For the RL agent, the manipulator state is given by \( s = (P_{\mathrm{S2R}}, g, \tilde{e}) \in \mathbb{R}^{18} \), which consists of the current pose after the S2R transfer \( P_{\mathrm{S2R}} \), the end-effector reaching goal \( g \), and the scaled pose error \( \tilde{e} = w \cdot (P_{\mathrm{S2R}} - g) \), where \( w \in \mathbb{R}^6 \) is a scaling vector used to balance the contribution of rotational and translational components. Notably, incorporating goal information into the state allows the policy to be trained under diverse goal conditions, resulting in a robust policy capable of reliable reaching across all positions within the workspace. The action is the sensor-space variables converted from $L_{\mathrm{spring}}$ in the simulation environment, i.e., $a=f_{\mathrm{sensor}}\in\mathbb{R}^{9}$. The reward function accounts for both goal completion and actuator saturation and is formulated as follows:    
\begin{equation}
\label{reward}
r = 
\begin{cases} 
R_g, & \text{if goal reached} \\
-\epsilon \cdot \|\tilde{e}\|_{2} -\zeta \cdot N, & \text{if goal unreached} \\
-R_s, & \text{if saturation occurs}
\end{cases}
\end{equation}

\noindent where a large positive reward \( R_g \) is given when the Euclidean norm of the scaled error \( \|\tilde{e}\|_2 \) falls below a threshold \( \theta \), indicating successful goal completion. Otherwise, a penalty is applied based on \( ||\tilde{e}||_{2} \), scaled by \( \epsilon \). The conducted environment steps \( N \) scaled by \( \zeta \) is also used to penalize unnecessary attempts and improve the exploration efficiency. Moreover, \( R_s \) is a large penalty applied when actuator saturation occurs.

The schematic flow of training the multi-goal RL algorithm in the simulation environment is shown in Fig.~\ref{fig:RLLoop}. At each step $t$, the current action \( a_{t} \) controls the soft manipulator to move from $P_{t}$ to $P_{t+1}$ in the parallelized simulation environment. Subsequently, the simulated end-effector poses \( (P_t, P_{t+1}) \) are first transformed into the adjusted practical pose \( (P_t, P_{t+1})_{\mathrm{S2R}} \) via the S2R transfer module. The current step transition \( (s_{t}, a_{t}, s_{t+1}, g) \) is then updated with the S2R-adjusted pose results before computing the reward \( r \). Finally, The transition, along with the corresponding action and reward \((s_{t}, a_{t}, s_{t+1}, g, r)\) is stored in the replay buffer for optimizing the control action $a_{t+1}$ for the next step. 

\subsubsection{Policy Optimization}
The Truncated Quantile Critics (TQC) is chosen for policy optimization as it stabilizes the learning process by truncating extreme quantiles of the value distribution (\cite{kuznetsov2020}), which mitigates overestimation bias in RL. With the combination of quantile truncation and entropy regularization, TQC balances exploration and exploitation while ensuring stable value estimates, making it ideal for high-dimensional tasks such as controlling redundant soft manipulator.

As highlighted in Fig.~\ref{fig:RLLoop}, during training, the TQC algorithm receives as input the transition tuple \( (s_t, a_t, s_{t+1}, g, r) \), which is sampled from the replay buffer. The critic network is updated using this full transition, while the actor network is trained to output the next action \( a_{t+1} \) based on the current state and goal. During deployment, only the trained actor network is used to generate actions conditioned on the current state and goal, enabling real-time policy execution without relying on critic evaluation or reward feedback.

Inspired by Hindsight Experience Replay (HER), we apply goal relabeling to quasi-static state transitions in RL framework to improve the learning efficiency. Given an RL trajectory consisting of a sequence of states and actions,
\begin{align}
    R = \{(s_0, a_0), \dots, (s_t, a_t),\dots, (s_T, a_T)\},
\end{align}
while the agent typically receives a reward \( R_{g} \) only when it reaches the goal \( g \). Subsequently, by relabeling each transition \( (s_t, a_t, s_{t+1}, g, r) \) as \( (s_t, a_t, s_{t+1}, g' =  P_{\mathrm{S2R}} \subseteq s_{t+1}, r = R_g) \), the end-effector pose of next state can be treated as a valid sub-goal that follows the same principle of HER. As a result, the replay buffer can be expressed as:
\begin{equation}
    R = \left\{ \left( s_t, a_t, s_{t+1}, g' = P_{\mathrm{S2R}} \subseteq s_{t+1}
 \right) \right\}_{t=0}^{T}.
\end{equation}

By adopting this relabeling approach, the agent can learn from previously unsuccessful transitions by treating intermediate states as goals, leading to improved sample efficiency, accelerated policy learning, and enhanced generalization in quasi-static goal sequence tracking. Furthermore, we implemented Stable-Baselines3 (SB3) using hyperparameters from the Contrib version to support policy training. Additionally, we adopted Gymnasium, a maintained fork of OpenAI’s Gym, for environment setup. Using SB3’s parallel processing with 14 workers, the framework significantly accelerates training, reducing training time while ensuring stable convergence. Each worker independently simulates agent interactions within the environment, updating a shared policy. This parallelization strategy improves exploration of the redundant action space while reducing computational bottlenecks.

%% file: Reward_Generation.tex
\section{GAIL-Based Reward Generation}\label{secIL}
\begin{figure}[t]
\centering
\includegraphics[width=\linewidth]{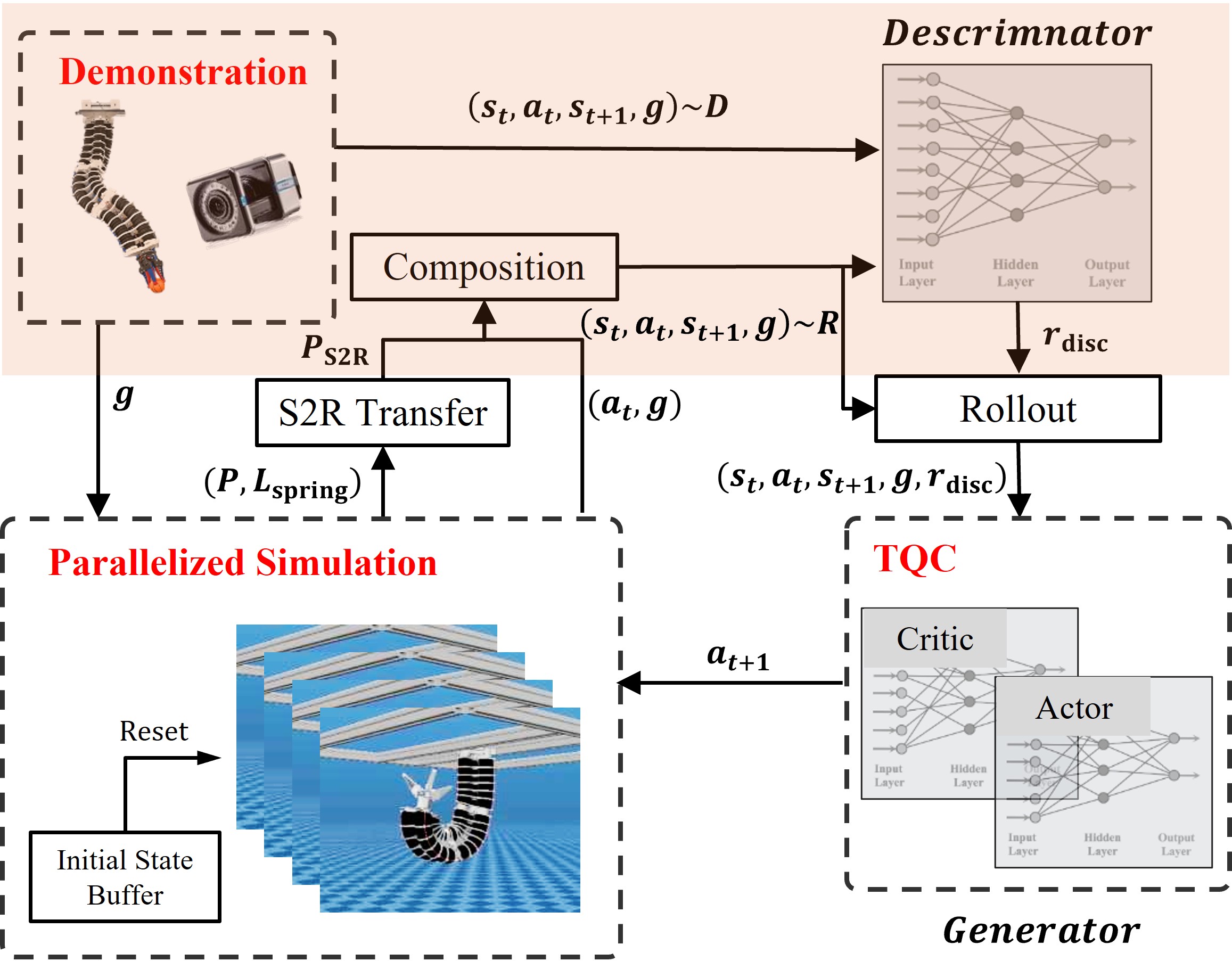}
\caption{Schematic flow of the SS-ILKC framework, which employs a GAIL-based approach to generate rewards from expert demonstrations. The newly added reward learning workflow, built upon our multi-goal RL algorithm in Sec.~\ref{secRL}, is highlighted in orange to enrich the learning from demonstrations in confined spaces.}\label{fig:ILLoop}
\end{figure}
For operations taken in confined spaces, the kinematic controller should not only ensure saturation-free execution but also account for complex environmental constraints in the configuration space. As a result, the feasible IK solution of soft manipulator goal-reaching is significantly restricted. Conventional RL fails to derive effective control policies for confined-space manipulation, as such behavior is difficult to be encoded through handcrafted rewards. In contrast, human operators can naturally demonstrate constraint-aware and time-efficient behavior when physically moving the soft manipulator. Therefore, we develop a new module into the SS-ILKC framework, which learns a reward function from expert demonstrations as illustrated in Fig.~\ref{fig:ILLoop}. Building upon our multi-goal RL framework, the GAIL-based reward generation constructs the reward signal by iteratively updating a discriminator to mimic expert behavior, thereby learning feasible kinematic control policies for confined-space operations.

\subsection{Definition of Sensor Space GAIL}
Given the challenges of manually designing a reward function for controlling soft manipulator kinematics in confined spaces, we introduce the SS-ILKC framework that leverages expert demonstrations to enable reward learning and accelerate control policy optimization. As shown in Fig. \ref{fig:ILLoop}, we collect a dataset of \( D \) expert sequences \( \{(s_0^j, a_0^j, s_1^j, \dots)\}_{j=1}^{D} \) from real-world demonstrations, where each sequence is generated by an expert attempting to reach a goal with the soft manipulator \( g^j \). Subsequently, a reward learning network, acting as a discriminator \( D_\psi \) parameterized by \( \psi \), is trained to distinguish between expert transitions \( (s, a,g) \sim D \) and those generated by the RL agent \( (s, a, g) \sim R \), where \( R \) represents the replay buffer used for training. The reward \( r_{\mathrm{disc}} \) is derived from the discriminator’s prediction which reflects the difficulty in distinguishing between expert and policy-generated samples. By maximizing \( r_{\mathrm{disc}} \), the RL agent is encouraged to generate control policies that align with expert demonstrations. 

During each training update, the simulator executes \( N \) environment steps and use an interval PD controller to track the desired sensor-space action $L_{\mathrm{spring}}$. To mitigate the simulation-to-reality gap, SS-ILKC also incorporates the S2R mechanism that transforms the end-effector pose in simulation \( P \) into the adjusted practical value \( P_{\mathrm{S2R}}\) to compose the state \( s \). Furthermore, the \( L_{\mathrm{spring}} \) in simulation is also converted into $f_{\mathrm{sensor}}$ to formulate the practical action \( a \). By integrating the S2R transfer within the RL agent, SS-ILKC ensures accurate and robust kinematic control of the soft manipulator while enabling zero-shot deployment without additional fine-tuning.

\subsection{Reward Learning Strategy}
\subsubsection{Reward Learning}
The overall workflow of learning the reward network and policy within the proposed SS-ILKC framework can be found in Fig.~\ref{fig:ILLoop}. In contrast to using the multi-goal RL-based framework with predefined reward functions, the reward is learned via an adversarial training process in SS-ILKC, where a discriminator \( D_\psi(s, a, g) \) is trained to distinguish between expert demonstrations \((s,a, g) \sim D\) and policy-generated samples \((s,a, g) \sim R\) by minimizing a loss function \(\mathcal{L}_{\mathrm{disc}}\) which will be specified later. The reward signal for the RL agent is then derived from the discriminator's output:
\begin{equation}
    r_{\mathrm{disc}}(s, a, g) = \log\big(1 - D_\psi(s, a, g)\big),
\end{equation}
\noindent where \((s,a, g) \sim R\), and is designed to encourage the agent to generate behavior that aligns with expert demonstrations.

\begin{figure}[t]
\centering
\includegraphics[width=\linewidth]{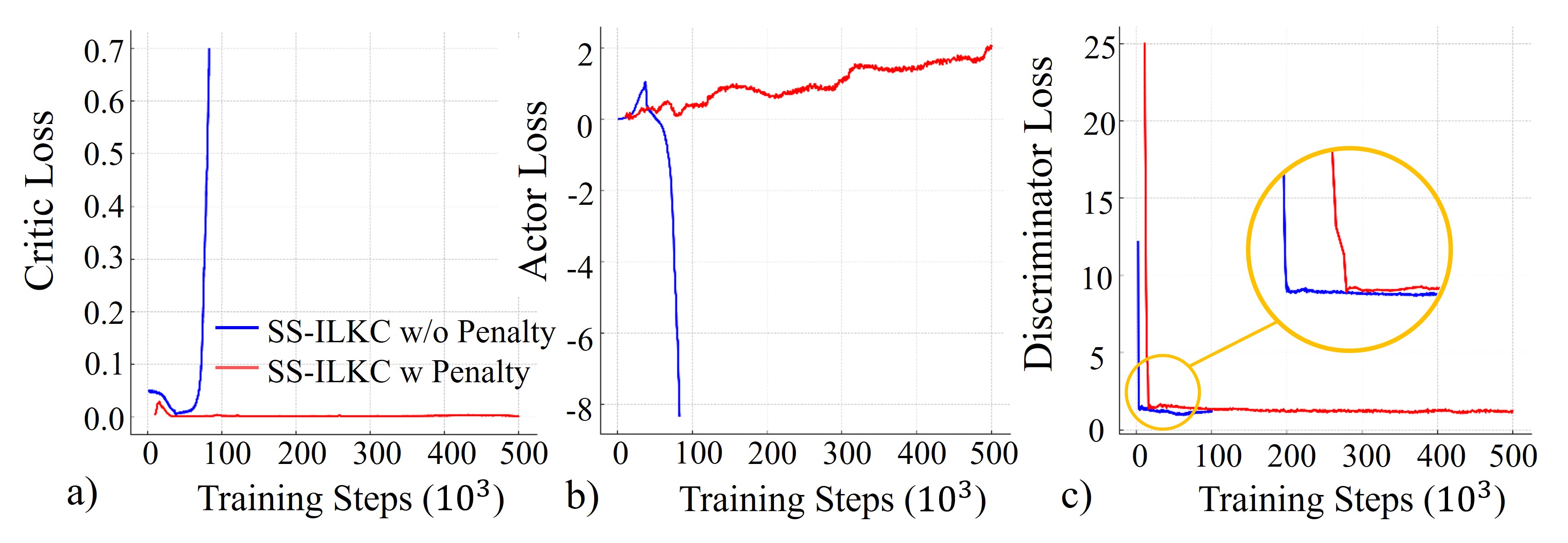}
\caption{Comparison of the losses with and without gradient penalty for: (a) critic loss, (b) actor loss, and (c) discriminator loss.}\label{fig:loss}
\end{figure}

\subsubsection{Generation of Demonstration Data}
Two widely used methods are commonly employed to demonstrate the robot manipulator's behavior: (1) remote control of the end-effector to reach the goal; (2) manual guidance of the end-effector through physical dragging. For soft manipulators, however, manual guidance can introduce lateral bending in individual chambers, which does not occur during actual pneumatic actuation. Therefore, to maintain consistency between simulation and real-world environments, we collect demonstrations via remote control. Due to the limited capability of existing simulators in handling interactive environmental constraints, such as confined-space operations, we opt to collect demonstrations directly in the real-world environment. In addition, the oversimplification of complex deformation, non-linear material properties, and uncertain dynamics of soft manipulator in simulation can lead to inaccuracies in conducting simulated manipulation. We argue that real-world demonstrations provide more accurate and physically consistent motion data, which supports reliable and transferable control policy learning.

\begin{algorithm}[t]
\caption{Discriminator Update Procedure}
\SetAlgoLined
\KwIn{Demonstrations \( \mathcal{D} = \{(s_t^j, a_t^j, s_{t+1}^j, g^j)\} \), replay buffer \( \mathcal{R} = \{\} \), policy \( \pi_\phi \), discriminator \( D_\psi \), gradient penalty weight \( \lambda \), batch size \( B \).}

\While{every update}{
    
    \# Sample goal uniformly from state space \\
    \( g \sim \text{Uniform}(s_t) \) \\

    \# Collect policy-generated transitions into replay buffer \\
    \( \mathcal{R} \gets \mathcal{R} \cup \{(s_t, a_t, s_{t+1}, g)\} \) sampled using \( \pi_\phi(g) \) \\

    \# Sample a batch from replay buffer \\
    \( \{s_t, a_t, s_{t+1}, g\} \sim \mathcal{R}, |B| \) \\

    \# Sample a batch from expert demonstrations \\
    \( \{s_t, a_t, s_{t+1}, g\} \sim \mathcal{D}, |B| \) \\

    \# Compute interpolated state-action pairs \\
    Compute \( \tilde{s}, \tilde{a} \) with \( \mu \sim \text{Uniform} (0, 1) \) \\

    \# Compute discriminator loss\\
    \( L_{\mathrm{disc}} \gets \log\big(1 - D_\psi(s_t, a_t, g)\big) + \log\big(D_\psi(s_t, a_t, g)\big) \) \\

    \# Compute gradient penalty term \\
    \( L_{\mathrm{gp}} \gets \lambda \cdot (\|\nabla_{\tilde{s}, \tilde{a}} D_\psi(\tilde{s}, \tilde{a}, g)\|_2 - 1)^2 \) \\

    \# Update discriminator parameters\\
    \( L_\psi \gets L_{\mathrm{disc}} + L_{\mathrm{gp}} \) \\
    \( \psi \gets \psi - \xi \cdot \nabla_\psi L_\psi \) \\

    \# Update rewards for policy optimization \\
    \( \forall (s_i, a_i, g_i) \in \text{batch}, r_i \gets \log\big(1 - D_\psi(s_i, a_i, g)\big) \) \\

    \# Reset replay buffer and update with new transitions \\
    \( \mathcal{R} \gets \emptyset \) \\
    \( \mathcal{R} \gets \mathcal{R} \cup \{(s_i, a_i, r_i, g_i)\} \) \\
}
\end{algorithm}

\subsubsection{Representation of Demonstration Data} 
We extend the relabeling strategy to expert demonstrations to improve sample efficiency in the SS-ILKC framework. Given a dataset of quasi-static demonstration sequences
\begin{equation}
    D = \{(s_0^j, a_0^j, \dots, s_t^j, a_t^j, \dots, s_T^j, a_T^j)\}_{j=1}^{D},
\end{equation}
every consecutive transition pair can be considered as a valid goal by treating the next state as the new goal, allowing the demonstration set to be rewritten as
\begin{equation}
    D = \left\{\left\{(s_t^j, a_t^j, g' =P_{\mathrm{S2R}} \subseteq s_{t+1}^j )\right\}_{t=0}^{T} \right\}_{j=1}^{D}.
\end{equation}

By integrating HER-inspired goal relabeling into both RL training and demonstration learning, we significantly enhance sample efficiency, improve policy robustness, and facilitate learning in quasi-static environments, while also making the reward function more generalizable for tracking similar paths.

\subsubsection{Loss Function and Training}
To ensure stable and efficient reward network training, we define the discriminator loss function as follows:
\begin{align}
\label{lossdis}
\mathcal{L}_{\mathrm{disc}} =& \mathbb{E}_{(s,a,g) \sim D}\big[\log\big(1 - D_{\psi}(s, a, g)\big)\big] \notag \\
&+ \mathbb{E}_{(s,a,g) \sim R}\big[\log\big(D_{\psi}(s,a,g)\big)\big] \notag \\
&+ \lambda \cdot \mathbb{E}_{\tilde{s}, \tilde{a} \sim \mathbb{P}_{\tilde{s}, \tilde{a}}} \big[ \|\nabla_{\tilde{s}, \tilde{a}} D_{\psi}(\tilde{s}, \tilde{a}, g)\|_2 - 1\big]^2.
\end{align}

This loss function consists of two components:
\begin{enumerate}
    \item \textbf{Discriminator Loss:} The first two terms quantify how well the discriminator differentiates expert transitions \( (s, a, g) \sim D \) from agent transitions \( (s, a, g) \sim R \). This follows the standard GAIL loss (ref.~\cite{ho2016}), encouraging the policy to mimic expert behavior.
    
    \item \textbf{Gradient Penalty:} The third term in \eqref{lossdis} stabilizes training by smoothing discriminator gradients (ref.~\cite{gulrajani2017}), reducing overfitting to sharp expert-policy differences.
\end{enumerate}
The values \( \tilde{s} \) and \( \tilde{a} \) are computed by linearly interpolating samples from the demonstration buffer \( D \) and the replay buffer \( R \), with a randomly sampled mixing coefficient \( \mu \in (0, 1) \) for each pair.




To evaluate the impact of the gradient penalty on the discriminator loss, we compared two conditions on training with and without the gradient penalty as shown in Fig.~\ref{fig:loss}. Results indicate that the discriminator loss converges rapidly without the gradient penalty and leads to instability in both the actor and critic losses. In contrast, incorporating the gradient penalty significantly improves stability, resulting in smoother and slower convergence. These findings demonstrate the effectiveness of the gradient penalty term for stabilizing training, reducing overfitting and enhancing the robustness of the discriminator. The pseudo code for updating $D_{\psi}$ has been given in ALGORITHM 1.

%% file: Experimental_Results.tex
\section{Experimental Results}\label{secResult}
\begin{figure}[t]
\centering
\includegraphics[width=\linewidth]{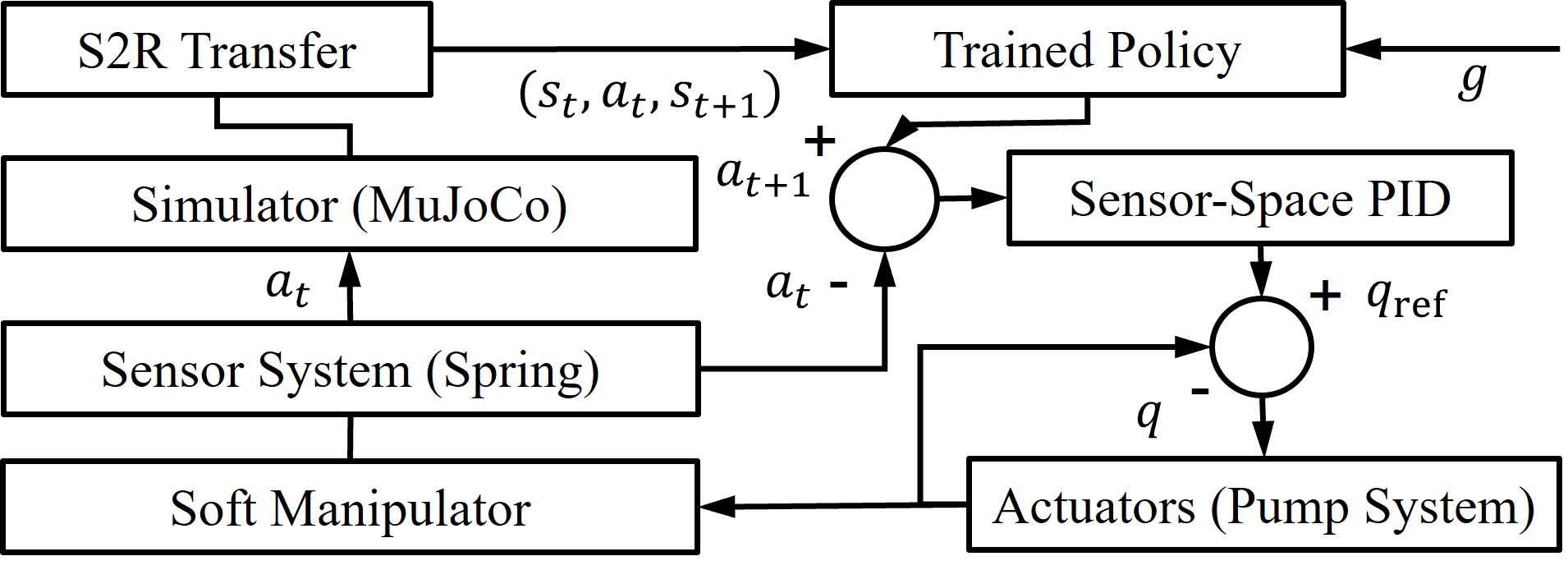}
\caption{The schematic diagram of deploying sensor-space policy to control the soft manipulator in reality. Given a goal $g$, the trained policy outputs the optimal control action, i.e., $a_{t+1}$ based on the current state-action set $(s_t,a_t,s_{t+1})$. $a_{t+1}$ will then serve as the control setpoint for a sensor-space Proportional–integral–derivative (PID) controller that computes the corresponding pressure values $q_{\mathrm{ref}}$ for the pump system to regulate its output $q$ for reaching the goal. }\label{fig:control}
\end{figure}

In this section, we validate the performance of the proposed SS-ILKC framework in achieving accurate and robust kinematic control of the soft manipulator under unknown loads and actuator saturation, and assess its generalization capability in both open and confined spaces. Notably, in all physical experiments, a motion capture system is only used for ground-truth verification but not integrated into the kinematic control process. The performance of our control method is also demonstrated in the supplementary video.

SS-ILKC is implemented on the soft manipulator shown in Figs.\ref{fig:problem3} and \ref{fig:workspace}. The system operates on a laptop with an Intel Core i9-11900K CPU and 32 GB of memory to enable real-time operation. The trained kinematic control policy network is deployed in a C++ program with a Qt-based GUI for user interaction and control. Fig.\ref{fig:control} presents the schematic diagram of deploying our SS-ILKC framework for the soft manipulator kinematic control.


\subsection{Path Following in Unloaded Conditions}\label{subsecFKValidation}
The first experiment assesses the soft manipulator’s path-following accuracy with a non-loads open-space scenario. The selected path is a 160 mm diameter circle with 40 uniformly distributed points along its path. The goal is to control the end-effector to precisely follow the given path. The implementation details of our SS-ILKC framework are detailed below.

\begin{figure}[t]
\includegraphics[width=\linewidth]{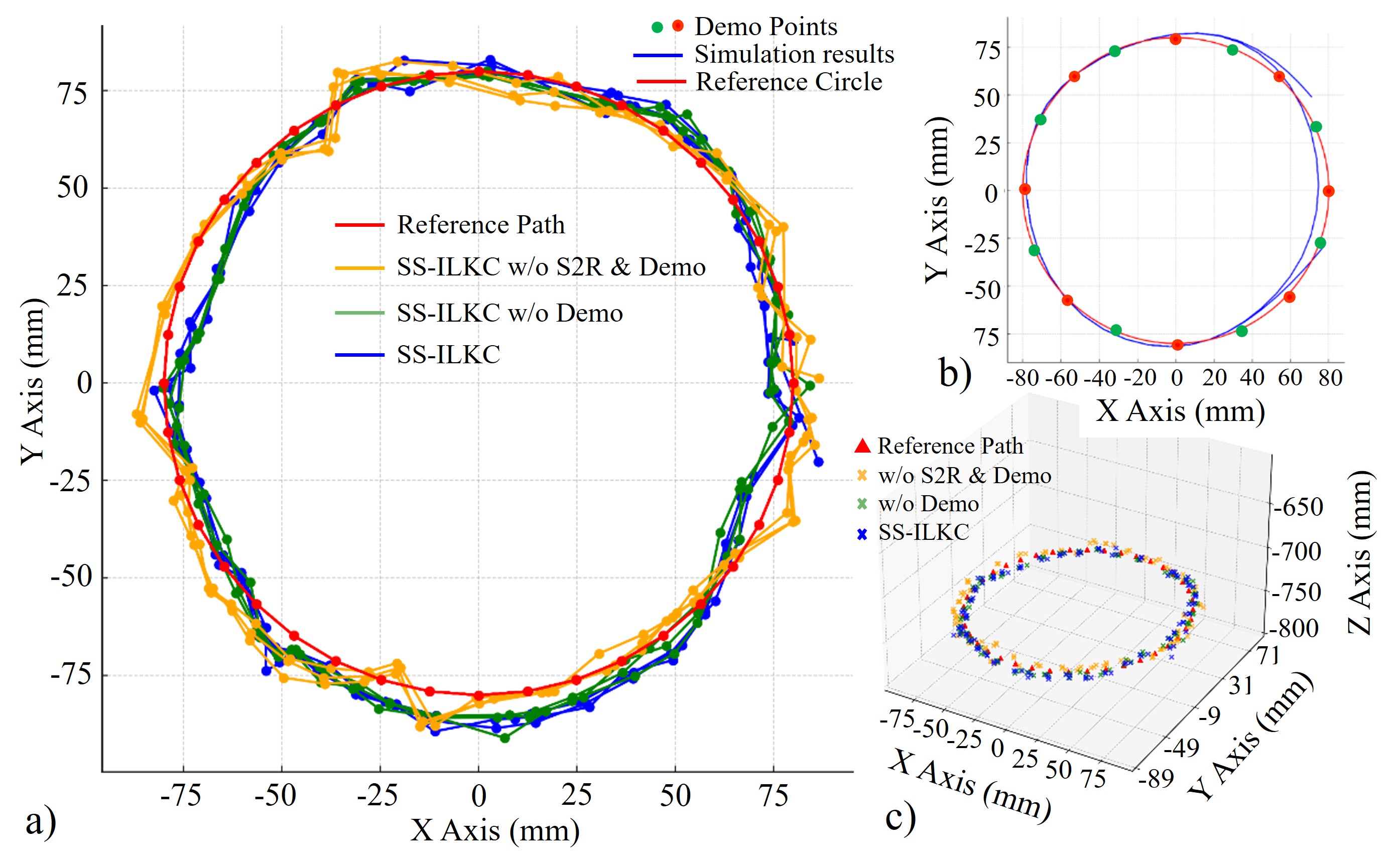} \vspace{2px} \\ 
\footnotesize
\begin{tabular}{l|cccc}
\hline \hline
\specialrule{0em}{2pt}{1pt}
                  & Trans. (mm)$^\dag$ & Yaw $(^\circ)$ & Pitch $(^\circ)$ & Roll $(^\circ)$ \\ 
                  \specialrule{0em}{1pt}{1pt}
                  \hline
                  \specialrule{0em}{1pt}{1pt}
                  		
w/o S2R \& Demo       &   21.01     & 13.72      &  2.82      &   6.98     \\
\specialrule{0em}{1pt}{1pt}
w/o Demo          &   11.59     & 9.54      &  2.44      &   2.99     \\
\specialrule{0em}{1pt}{1pt}
SS-ILKC         &   11.29     & 
9.98      &  1.88      &   1.60     \\ 
\specialrule{0em}{1pt}{1pt}
\hline \hline
\end{tabular}
\begin{flushleft}
$^\dag$Translation errors are computed by $\sqrt{\Delta x^2 + \Delta y^2 + \Delta z^2}$.
\end{flushleft}
\caption{Control performance of path-following using our SS-ILKC framework under three different conditions: (1) w/o S2R transfer and physical demonstration, (2) with S2R transfer but w/o demonstration, and (3) full deployment. (a) 2D view of the reference and ground-truth paths under different deployment. (b) The selected demonstration points and the soft manipulator's path were obtained in the physical simulator without applying S2R transfer. (c) A 3D view of the path-following results for different approaches. The table below summarizes the path tracking accuracy by evaluating the root mean square error of translation in the X-, Y- and Z-coordinates, as well as the average errors in rotation which is decomposed into Yaw, Pitch, and Roll.
}\label{fig:Circle_analysis}
\end{figure}

\begin{enumerate} 
\item Demonstration data collection: The demonstration dataset consists of 16 uniformly distributed points along the path, recorded using the motion capture system.

\item Initial training with partial demonstration data: SS-ILKC was first trained on a subset of 8 selected points (highlighted in red in Fig.~\ref{fig:Circle_analysis}(b)) for $1 \times 10^5$ steps. This initial phase enables the reward network to approximate the desired goal while mitigating the risk of catastrophic forgetting in later training stages.

\item Training on full demonstration data: SS-ILKC was subsequently trained with all 16 points in Fig.~\ref{fig:Circle_analysis}(b) for an additional $2 \times 10^5$ steps. During this phase, the replay buffer was balanced by retaining 20\% of the initial training samples to ensure equal representation of each goal, further refining the policy for accurate path following.
\end{enumerate}

We evaluated the performance of SS-ILKC and conducted ablation studies using two other variants: (1) without S2R transfer and physical demonstration and (2) with S2R transfer but without physical demonstration. The scaling vector of pose error is defined as $w=\mathrm{diag}([0.0056\times \textbf{I}^{3}, 0.001 \times \textbf{I}^{3}])$, the parameters in the reward function (\ref{reward}) are set as \( R_g = 100 \), \( \epsilon = 10 \), \( \zeta = 0.1 \) and \( R_s = 100 \), and the goal-reaching threshold is defined as \( \theta = 0.03 \). The gradient penalty weight $\lambda$ in GAIL loss (\eqref{lossdis}) is set to \( \lambda = 20 \). The quantitative results for the path-following task are shown in Fig.~\ref{fig:Circle_analysis}. SS-ILKC achieved the best path-tracking performance, with an average translation tracking error of \SI{11.29}{mm}. SS-ILKC without physical demonstration (with S2R transfer) also demonstrated competitive performance, yielding an average translation error of \SI{11.59}{mm}. In contrast, learning policy without physical demonstration and S2R transfer resulted in a significantly higher translation error of \SI{21.01}{mm}, nearly twice that of the other two variants.
Although the path lies on a planar surface, maintaining a stable vertical orientation throughout the motion introduces additional complexity. In particular, the top two modules of the soft manipulator must actively control five DoFs to complete the task. Throughout the entire path, the average errors for Yaw, Pitch, and Roll were \SI{9.98}{\degree}, \SI{1.88}{\degree}, and \SI{1.60}{\degree}, respectively. Notably, without S2R transfer and physical demonstration, the learning framework exhibited a training error of \SI{3.89}{mm} in the simulation environment, as shown in Fig.~\ref{fig:Circle_analysis}(b), indicating that the primary source of error stems from the mismatch between the simulation and real-world system.

In summary, these results demonstrate the effectiveness of the SS-ILKC framework in path following tasks, showcasing its ability to achieve precise kinematic control in open space under non-loading conditions. Furthermore, leveraging multi-goal policy optimization, the fully deployed SS-ILKC and the variant only with S2R transfer both exhibit strong generalization capabilities, successfully tracking a denser set of points on the selected path (trained on 16 points, executed on 40 points). The degraded control performance without S2R transfer underscores the importance of the proposed S2R transfer mechanism in bridging the gap between simulation-trained policies and practical deployment.

\subsection{Path Following in the Presence of Unknown Loads}\label{subsecPathWithUnknownLoad}
\begin{figure}[t]
\centering
\includegraphics[width=\linewidth]{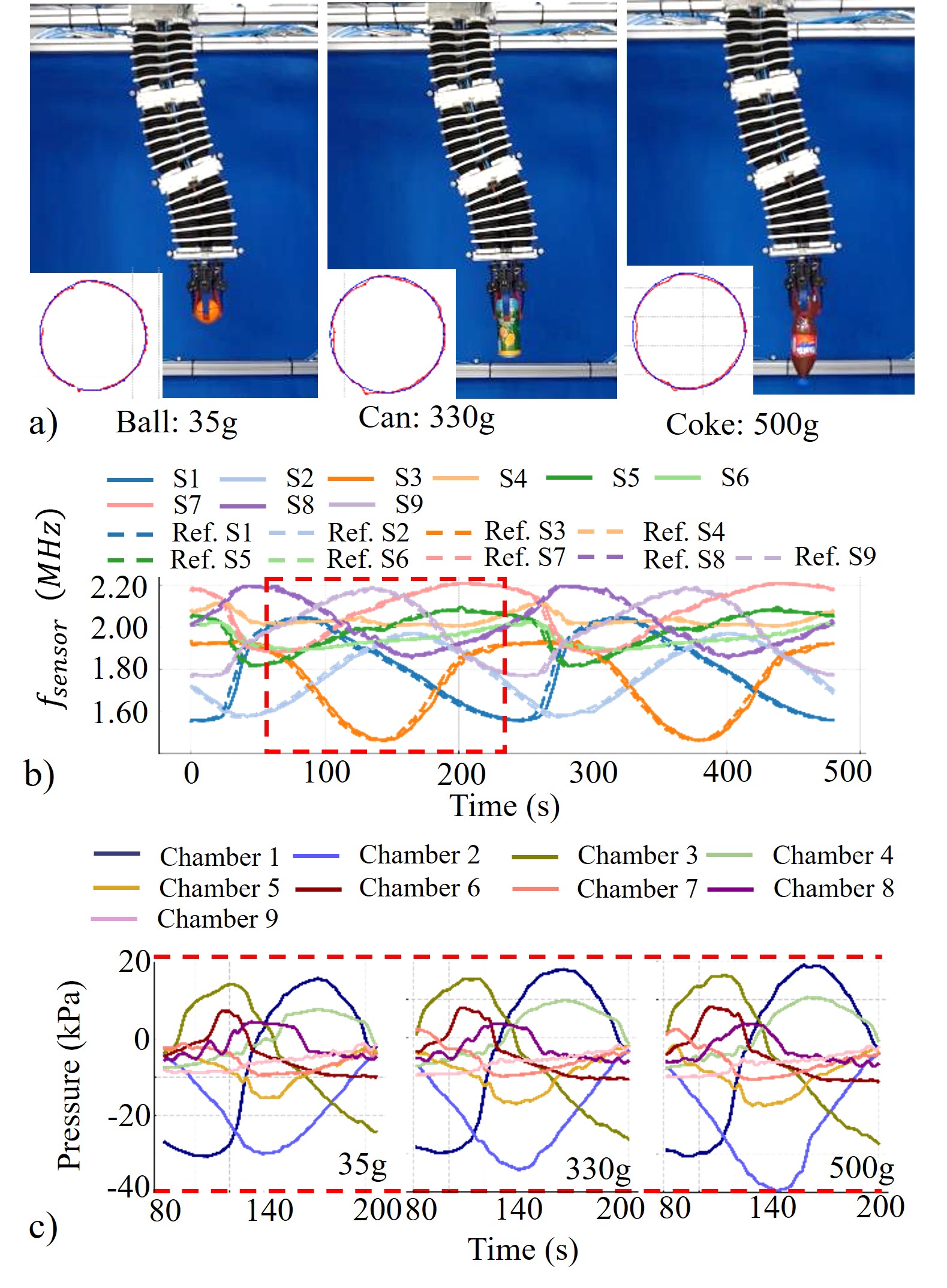}
\caption{Performance of SS-ILKC in the path following task in the presence of different loads. (a) The results of holding objects with weights as \SI{35}{g}, \SI{330}{g} and \SI{500}{g}, where blue and red circles represent reference and actual manipulator's paths respectively. (b) The reference and actual sensor feedback while holding the ball. (c) Pressure in chambers during \SI{80}{s} to \SI{200}{s} with different loads where chamber 2 is close to a saturation (dashed red line) due to the increased load.}\label{fig:circle_different_weight}
\end{figure}

\begin{figure*}[t]
\includegraphics[width=\linewidth]{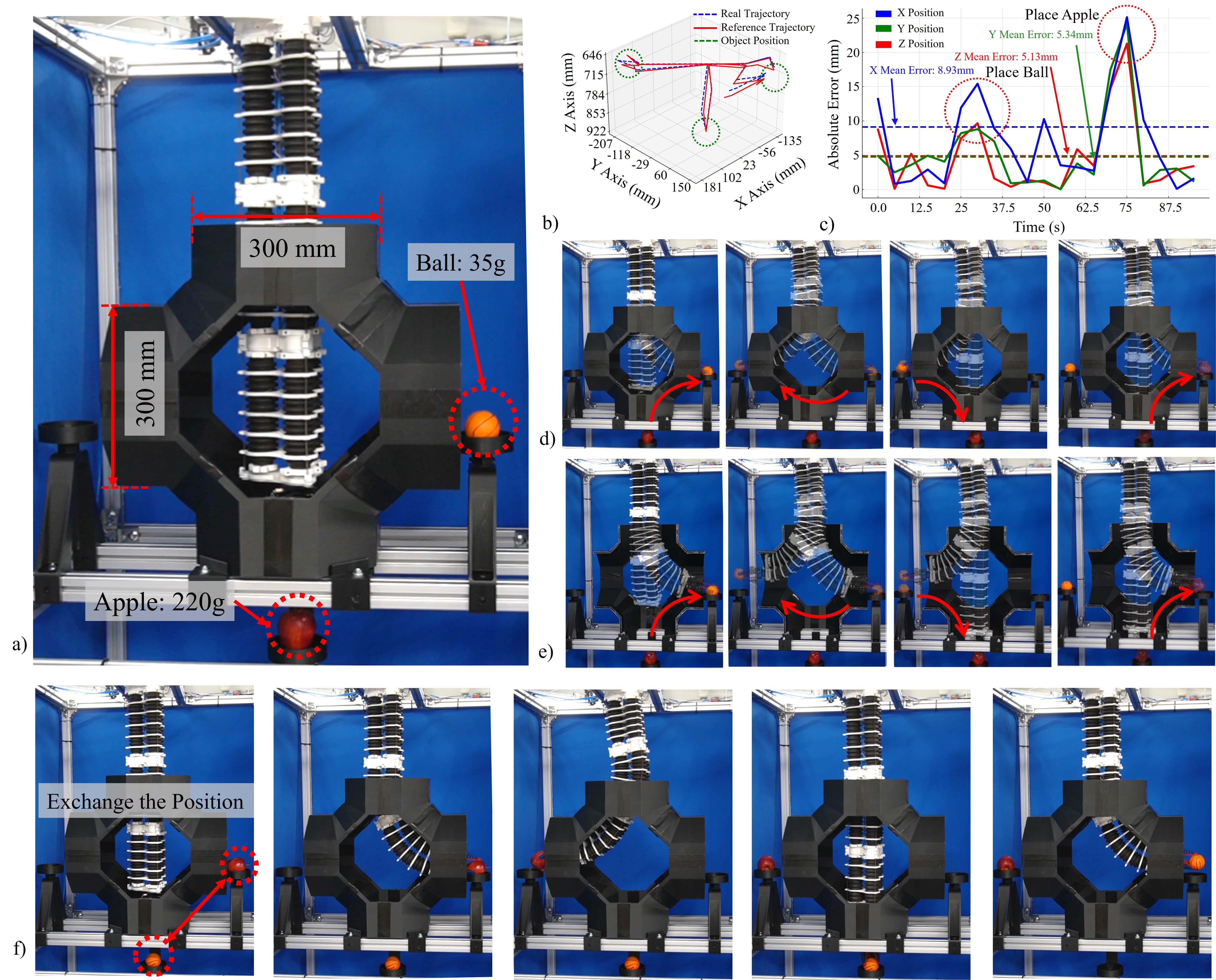}
\caption{Performance verification of our SS-ILKC framework in executing a series of pick-and-place tasks within the confined space of a cross-shaped pipe. (a) Experimental setup with the specified dimensions and the weights of objects. (b) Demonstrated path, SS-ILKC path, and the goals as positions for the pick-and-place tasks. 
(c) Absolute (solid lines) and mean (dashed lines) errors between the demonstrated and realized paths in X-, Y-, and Z-coordinates. 
(d) Completion of two pick-and-place tasks with different loads and paths. 
(e) Visualization of soft manipulator's states by replacing the fully covered pipe with a half-displayed pipe. 
(f) Results after exchanging the positions of two objects and executing the same task using the same SS-ILKC policy trained from the demonstrations.
}\label{fig:pick_place_blk}
\end{figure*}

In addition to evaluating circular path following tasks in unloaded conditions, we further validated the loading independence of our SS-ILKC framework by applying unknown loads to the gripper at the end-effector. To ensure a fair comparison, we employed the same trained control policy as in Sec.~\ref{subsecFKValidation} and tested five objects with different weights: a sponge ball (\SI{35}{g}), a tape (\SI{105}{g}), an apple (\SI{220}{g}), a beverage can (\SI{330}{g}), and a bottle of Coke (\SI{500}{g}). 

Three representative objects were selected for evaluation, and the experimental results are illustrated in Fig. \ref{fig:circle_different_weight}, where the operational snapshots and corresponding path following plots (blue and red lines represent reference and actual manipulator path respectively) for different load conditions are shown in Fig. \ref{fig:circle_different_weight}(a). The results indicate that, despite variations in the applied loads, the soft manipulator successfully completes the path-following task. Notably, the average translation error across five different objects is approximately \SI{9}{mm}, demonstrating good path following performance comparable to that observed under non-loading conditions. Furthermore, Fig. \ref{fig:circle_different_weight}(b) presents the sensor feedback $f_{\mathrm{sensor}}$ values during the path following process while holding the ball, including the reference sensor values generated by the learned policy and the sensor values recorded by the sensor board. The close alignment between the two indicates that the reference points provided by the trained policy are consistently feasible and achievable for the soft manipulator. Additionally, Fig. \ref{fig:circle_different_weight}(c) illustrates the variation in pressure values over a specific period, highlighted by the red dashed rectangle in Fig. \ref{fig:circle_different_weight}(b). These variations correspond to different load conditions, as shown in Fig. \ref{fig:circle_different_weight}, allowing us to assess actuation characteristics under varying loads. It is evident that, as the load increases, the pressure required to realize the same reference spring length also increases. Despite such variations, the soft manipulator maintains its control performance while dynamically adjusting the pressure to compensate for the load and ensuring an accurate path tracking.

In summary, experimental results of the path following tasks confirm the robustness of our SS-ILKC framework in handling unknown loads. The load-independent control policy, trained without external loads, adapts to varying loads, demonstrating the adaptability and generalization of our approach. 

\subsection{Pick-and-Place in Confined Scenarios}\label{subsecthreeChamberActuator}
To evaluate the efficiency of the proposed SS-ILKC framework for soft manipulator operation in confined spaces, we perform a kinematic control task requiring object pick-and-place within a cross-shaped pipe, as shown in Fig.~\ref{fig:pick_place_blk}(a). The pipe has a diameter of 300 mm, where the soft manipulator is initially positioned through the top outlet. Three containers are placed at the left, right, and bottom outlets. The soft manipulator is tasked with performing the following pick-and-place operations:
\begin{enumerate}
\item Pick up a sponge ball (\SI{35}{g}) from the right container;
\item Place the sponge ball onto the left container;
\item Pick up an apple (\SI{220}{g}) from the bottom container;
\item Place the apple onto the right container.
\end{enumerate}

Similar to the implementation details summarized in Sec.~\ref{subsecFKValidation}, a human operator remotely controlled the soft manipulator to collect demonstration data for policy training. Using a motion capture system, 28 demonstration positions were recorded for reward learning, enabling the soft manipulator to operate within a confined space. These positions included key waypoints essential for task completion, with four terminal pick-and-place states illustrated in Fig.~\ref{fig:pick_place_blk}. SS-ILKC was trained for a total of $6 \times 10^5$ steps using the same balanced training strategy described in Sec.~\ref{subsecPathWithUnknownLoad}.

To optimize the learning process and mitigate catastrophic forgetting, 
we adjusted the proportion of each demonstration in the replay buffer every \( 1 \times 10^5 \) steps on purpose. 
These adjustments were guided by the manipulator's training performance, prioritizing underperforming demonstrations while ensuring adequate coverage of all examples.

During experiments, We first compare the demonstration path with the practical path of soft manipulator as shown in Fig.~\ref{fig:pick_place_blk}(b), where the manipulator closely follows the demonstration path using the control policy trained by our SS-ILKC framework. Additionally, Fig.~\ref{fig:pick_place_blk}(c) presents the absolute and mean errors along the X-, Y-, and Z-coordinates for each demonstration position. The average translation errors between the executed and demonstrated positions are \SI{8.93}{mm}, \SI{5.34}{mm}, and \SI{5.13}{mm}, respectively, demonstrating the soft manipulator’s accuracy in replicating the demonstrated motions. Noted that two peak errors, highlighted by dashed red circles in Fig.~\ref{fig:pick_place_blk}(c), correspond to key waypoints where the manipulator places objects into different containers. These errors primarily result from slight contact between the gripper and the container during placement. Furthermore, as the ball or apple is transferred to the container, the sudden change in loading conditions introduces a temporary tracking error. However, SS-ILKC can effectively mitigate this error within a short period, as shown in Fig.~\ref{fig:pick_place_blk}(c).

To illustrate the detailed motions of the soft manipulator during the pick-and-place task within the pipe, Fig.~\ref{fig:pick_place_blk}(d) and (e) show the soft manipulator's states at four key waypoints during pick-and-place with a fully covered and half-displayed pipe, respectively. The results indicate that the soft manipulator successfully replicates the demonstrated motions, operating smoothly within the confined pipe. Furthermore, to validate the robustness of SS-ILKC in handling unknown loads, we apply the same policy network while exchanging the positions of the apple and the ball. As shown in Fig.~\ref{fig:pick_place_blk}(f), despite the changes in load across different tasks, the soft manipulator successfully completes the task using the same policy network. This demonstrates the robustness of our sensor-space-based learning framework in adapting to varying and unknown loads without requiring additional retraining.


\begin{figure}[t]
\centering
\includegraphics[width=\linewidth]{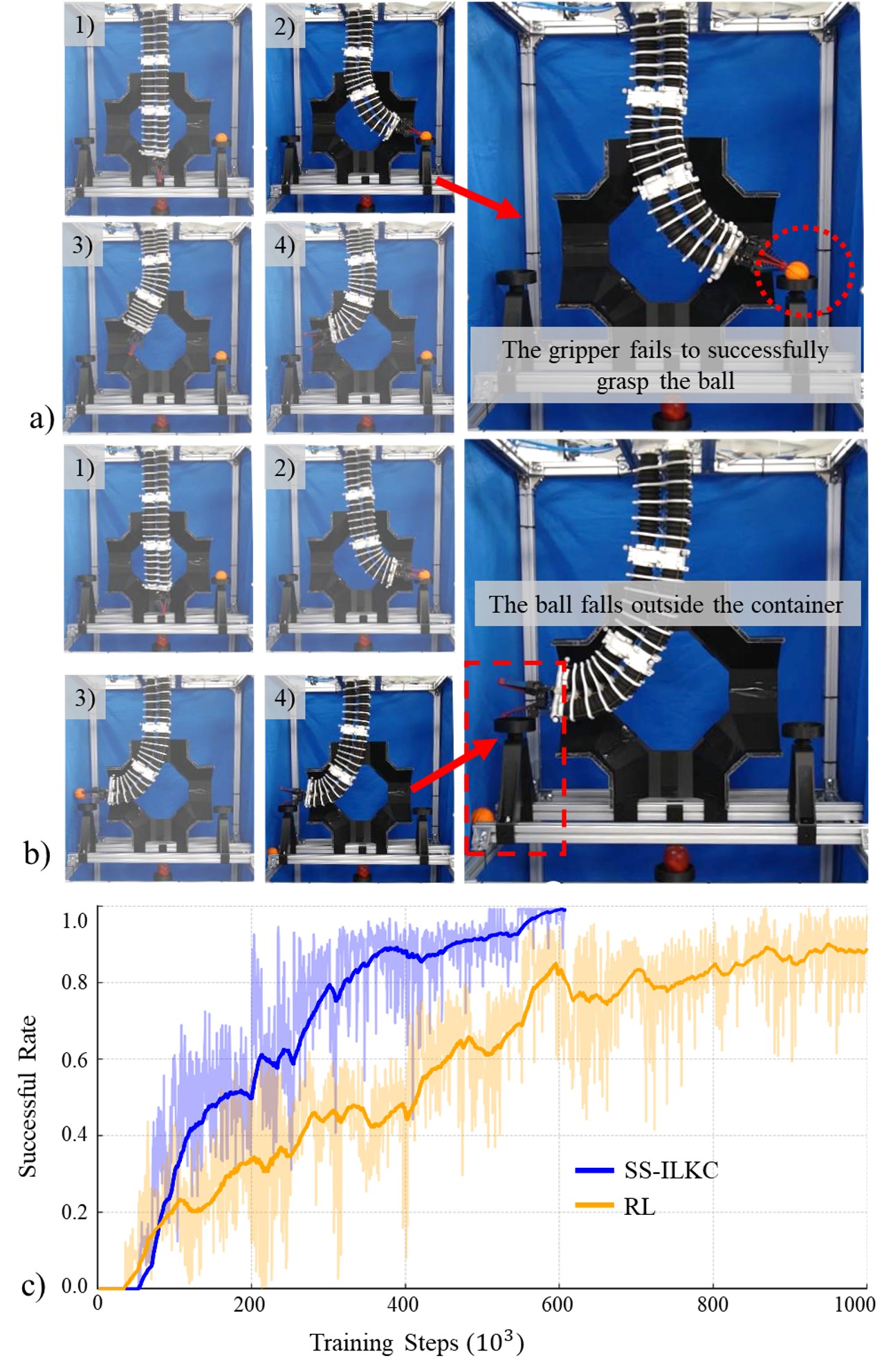}
\caption{Performance comparison of SS-ILKC with and without physical demonstration. (a) Task executions of SS-ILKC trained without physical demonstration after \(3 \times 10^5\) steps -- the gripper fails to grasp the ball from the right container. (b) SS-ILKC trained without physical demonstration after \(6 \times 10^5\) steps -- the gripper successfully grasps the ball but fails to drop it into the left container. (c) Comparing to the full SS-ILKC framework trained with physical demonstrations, the successful rates of SS-ILKC without physical demonstration over training steps are much lower in general.
}\label{fig:Failure}
\end{figure}

\begin{figure*}[t]
\includegraphics[width=\linewidth]{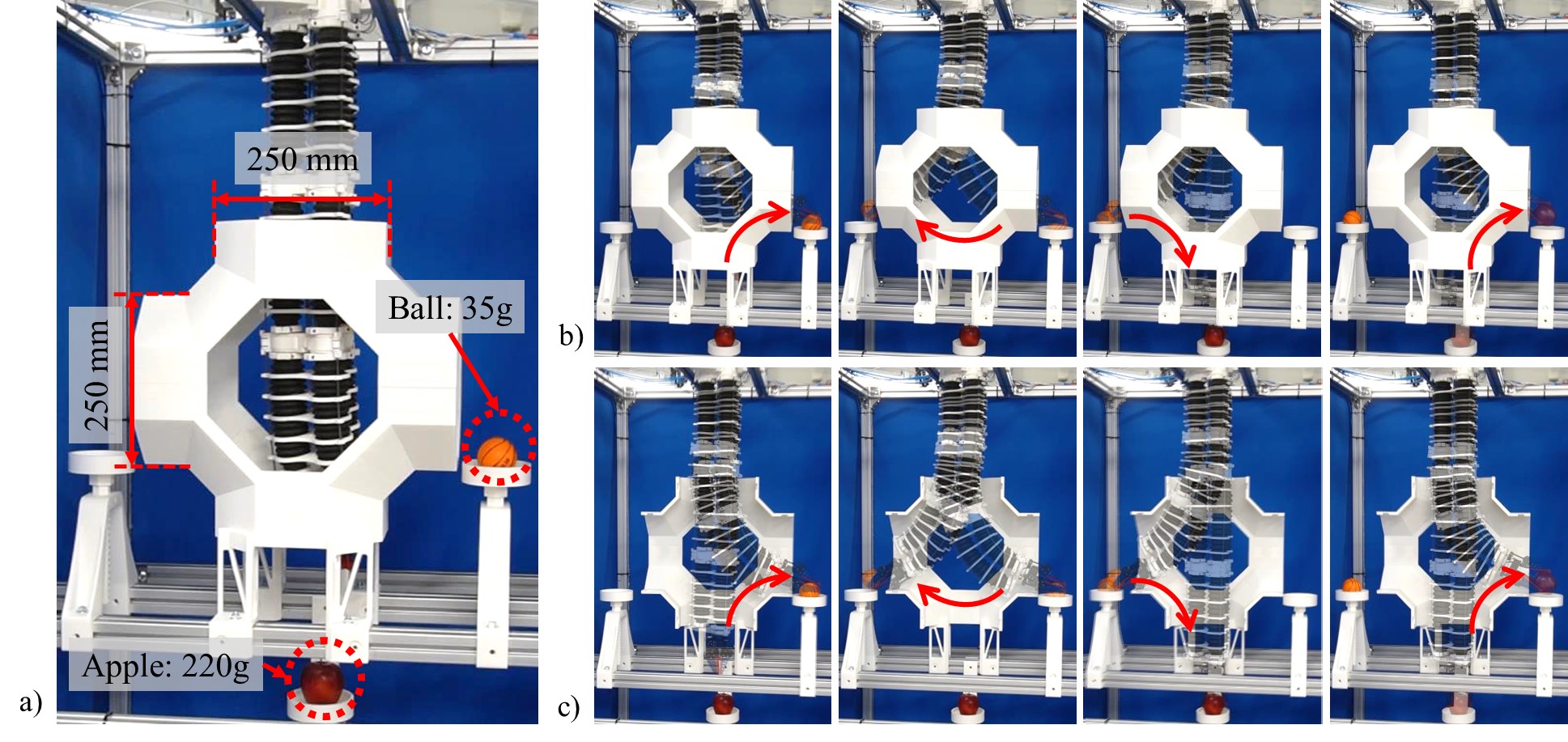}
\caption{Generalizability of the trained reward network -- in these pick-and-place tests, we apply the policies trained on a setup as shown in Fig.\ref{fig:pick_place_blk} on a new setup as shown in (a) with the pipe diameter reduced from \SI{300}{mm} to \SI{250}{mm}. (b) Completion of two pick-and-place tasks with different loads and paths. (c) Visualization of soft manipulator's states by replacing the fully covered pipe with a half-displayed pipe.
}\label{fig:pick_place_white}
\end{figure*}

To provide a comparison, we also conducted an ablation study by training the SS-ILKC only with the reward function (i.e., \eqref{reward}) for \(6 \times 10^5\) steps. The same training was performed but alternatively using the demonstration points as training goals. The experimental results are illustrated in Fig.~\ref{fig:Failure}. In Fig.~\ref{fig:Failure}(a), we first evaluate the soft manipulator's behavior after \(3 \times 10^5\) steps. At this stage, the soft manipulator failed to precisely reach the target position and grasp the ball from the right container. As training progressed to \(6 \times 10^5\) steps—the same number of steps at which SS-ILKC successfully completed all tasks—the policy trained without physical demonstration was able to grasp the ball. However, during the drop-off phase, the ball fell outside the left container, as depicted in Fig.~\ref{fig:Failure}(b). We then train this failure policy for an additional \(4 \times 10^5\) steps and the successful rates of two methods (\(1 \times 10^6\) in total for RL) are presented in Fig.~\ref{fig:Failure}(c). Initially, SS-ILKC trained with GAIL exhibited a slower success rate than using reward function due to the added complexity of the reward network during early training. However, as training progressed, SS-ILKC trained with GAIL achieved faster convergence and higher success rates. Without learning proper reward from physical demonstration, although the control policy's success rate improved over time, policy trained with reward function required significantly more steps and still could not surpass GAIL approach. These results highlight that the ability to learn from demonstrations enables SS-ILKC to facilitate confined-space operations while requiring fewer computational resources and less training time

Moreover, we evaluate the generalizability of SS-ILKC by deploying the reward network, trained on a black pipe with a diameter of \SI{300}{mm}, to task executions in a white pipe with a smaller diameter of \SI{250}{mm}. The reduced diameter introduces tighter spatial constraints, making the pick-and-place task more challenging. This task involves 37 goal positions for multi-goal RL policy optimization, comparable to the black pipe scenario. Using only the new target goals, the RL agent is trained for $6 \times 10^5$ steps with the previously learned reward network to perform the pick-and-place tasks in the new pipe. The experimental results, shown in Fig.~\ref{fig:pick_place_white}, demonstrate that the soft manipulator successfully completes a series of pick-and-place tasks. These findings highlight the strong generalization capabilities and training efficiency of the our SS-ILKC policy learned from physical demonstration, even under more constrained conditions.

%% file: Conclusion.tex
\section{Conclusion and Discussion}\label{secConclusion}
In this paper, we propose the SS-ILKC framework as a learning based method for robust kinematic control of a redundant soft manipulator under unknown loads, actuator saturation, and even operating in confined-space. By leveraging expert demonstrations, the sensor-space GAIL-based approach enables efficient reward learning, guiding a multi-goal RL optimizer to derive the control policy for accurate path following in both open and confined environments. The framework integrates a physical simulator adapted for soft manipulators and incorporates a sim-to-real transfer mechanism to minimize the gap between simulation and real-world deployment. Experimental evaluations have conducted to demonstrate the effectiveness of SS-ILKC, achieving saturation-free manipulation of objects with unknown weights, robust path following, and generalization beyond the training demonstrations.

\begin{figure}[t]
\centering
\includegraphics[width=\linewidth]{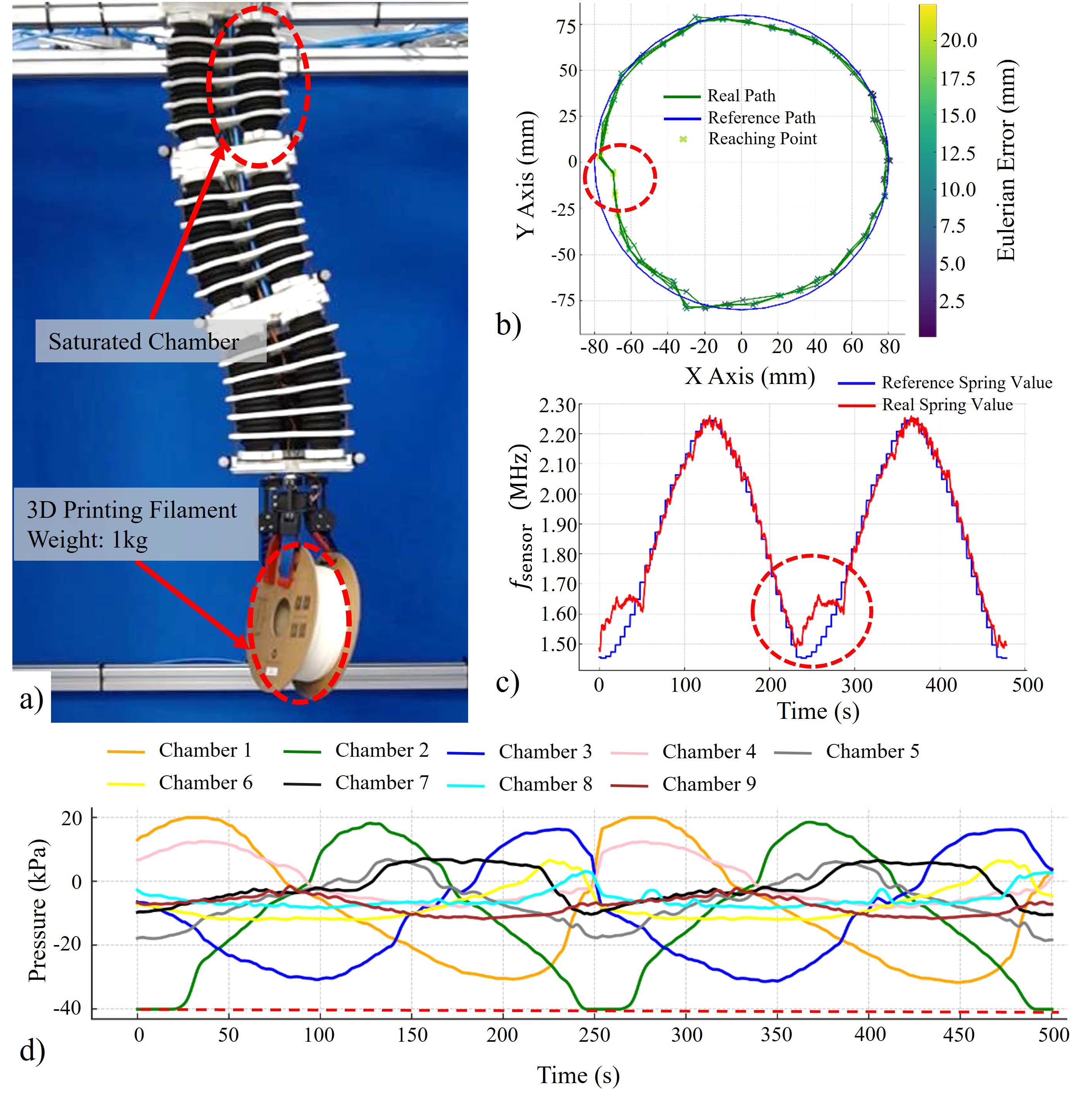}
\caption{A limitation of our approach. (a) Chamber saturation occurs at the top section of the soft manipulator when a very large load is applied (e.g., \SI{1}{kg} in this case). (b) The top view of path following results illustrate where the saturation happens. (c) Reference and actual sensor values during the path following task, and (d) pressures inside chambers.}\label{fig:filament}
\end{figure}

Despite achieving high path-following accuracy, SS-ILKC has certain limitations, with control saturation under relatively large loads being a primary concern. As shown in Fig.\ref{fig:filament}, a \SI{1}{kg} load was attached to the soft manipulator while performing the same circular path described in Sec.\ref{subsecFKValidation}. A significant deviation, marked by the red circle in Fig.\ref{fig:filament}(b), resulted in a maximum tracking error of \SI{23.76}{mm}. Further analysis in Figs.\ref{fig:filament}(c) and (d) indicates that the sensor feedback from chamber 2 deviated substantially from its reference values, causing actuator saturation. Consequently, the chamber failed to achieve the desired spring length, preventing the manipulator from successfully completing the target path. This issue is likely attributed to calibration errors near the workspace boundary. Future work should focus on developing more robust sampling methods to accurately characterize input saturation within the null space (ref.~\cite{cao2021reinforcement,izadbakhsh2020robust}). Additionally, a comprehensive analysis of the workspace under varying loading conditions (ref.~\cite{amehri2022fem}) could further enhance the consistency and reliability of kinematic control performance. 

Another limitation of the proposed framework is the increased training data requirements and computational costs when smoother movements are desired, as more static state data must be collected. This also restricts the use of advanced RL training techniques, such as HER (\cite{andrychowicz2017hindsight}), which could otherwise improve learning efficiency. To address this issue, future study could explore explicit modeling of pneumatic actuator dynamics (\cite{xavier2021design,joshi2021pneumatic}) and co-optimization of task and sensor placement (\cite{spielberg2021co}) to enable smoother and more cost-effective control of soft manipulators.

Furthermore, while SS-ILKC has successfully demonstrated its ability to operate in tightly confined spaces, the spatial location of the pipe is predefined in the current implementation. For dynamically evolving confined environments (e.g., the workspace changes over time or having dynamic obstacles), more advanced approach needs to be developed by integrating advanced perception solutions (\cite{shih2020electronic}) and interactive modeling of unknown environments (\cite{della2020model}).